\documentclass[journal]{IEEEtran}

\usepackage{graphicx}
\usepackage{caption}
\usepackage{subcaption} %
\usepackage{xcolor} %
\usepackage{amsmath} %
\usepackage{amssymb}

\usepackage{colortbl}
\usepackage{booktabs}
\usepackage{tabularx}
\usepackage{multirow}

\usepackage{amsmath}
\usepackage{siunitx}
\usepackage[acronym,nomain,nonumberlist]{glossaries}
\newacronym{lidar}{LiDAR}{Detection and Ranging}
\newacronym{slam}{SLAM}{Simultaneous Localication and Mapping}

\makeglossaries

\newif\ifready
\readytrue
\newif\ifannoy
\DeclareMathOperator*{\argmax}{arg\,max}

\makeatletter
\def\blfootnote{\gdef\@thefnmark{}\@footnotetext}
\let\NAT@parse\undefined
\makeatother
\usepackage[hidelinks]{hyperref} %
\usepackage[nameinlink,capitalise]{cleveref} %
\usepackage{url}
\usepackage{soul}

\usepackage[export]{adjustbox}

\definecolor{TableBlue}{rgb}{0.17,0.49,0.75}
\definecolor{Cerulean}{rgb}{0,0,0.95}
\definecolor{LimeGreen}{rgb}{0.15,0.65,0.15}
\definecolor{RoyalBlue}{rgb}{0.18,0.28,0.92}
\definecolor{Rose}{rgb}{1.0, 0.15, 0.21}
\definecolor{Orange}{rgb}{1.0, 0.5, 0.0}
\definecolor{Gray}{gray}{0.6}
\definecolor{Black}{gray}{0.0}
\definecolor{Purple}{rgb}{0.77,0.12,0.64}
\definecolor{FullBlue}{rgb}{0,0,1}

\usepackage{orcidlink}

\begin{document}
\title{\LARGE \bf 
HiMo: High-Speed Objects Motion Compensation in Point Clouds
}

\author{
Qingwen~Zhang\orcidlink{0000-0002-7882-948X},~\IEEEmembership{Graduate Student Member, IEEE}, 
Ajinkya~Khoche\orcidlink{0009-0009-6935-6797},~\IEEEmembership{Member, IEEE},
\\
Yi~Yang\orcidlink{0000-0002-6679-4021},~\IEEEmembership{Graduate Student Member, IEEE},
Li~Ling\orcidlink{0000-0001-7687-3025},~\IEEEmembership{Student Member, IEEE},
\\
Sina~Sharif~Mansouri\orcidlink{0000-0001-7631-002X},
Olov~Andersson\orcidlink{0000-0001-7248-1112},~\IEEEmembership{Member, IEEE},
Patric~Jensfelt\orcidlink{0000-0002-1170-7162},~\IEEEmembership{Member, IEEE}

\thanks{
Received 28 August 2024; revised 27 April 2025; accepted 19 August 2025. 
Date of publication 8 October 2025; date of current version 22 October 2025.
This work was partially supported by the Wallenberg AI, Autonomous Systems and Software Program (WASP) funded by the Knut and Alice Wallenberg Foundation and Prosense (2020-02963) funded by Vinnova. 
This paper was recommended for publication by Associate Editor M.~Walter and Editor S.~Behnke upon evaluation of the reviewers' comments. \textit{(Corresponding author: Qingwen~Zhang.)}
}
\thanks{Qingwen~Zhang, Li~Ling, Olov~Andersson, and Patric~Jensfelt are with the Division of Robotics, Perception, and Learning, KTH Royal Institute of Technology, Stockholm 114 28, Sweden. (email: qingwen@kth.se)}
\thanks{Ajinkya~Khoche, Yi~Yang are with the Division of Robotics, Perception, and Learning, KTH Royal Institute of Technology, Stockholm 114 28, Sweden, and also with Scania Group, Södertälje 151 87, Sweden.}
\thanks{Sina~Sharif~Mansouri is with Autonomous Transport Solutions Lab, Scania Group, Södertälje 151 87, Sweden.}
\thanks{Digital Object Identifier 10.1109/TRO.2025.3619042.}
}
\markboth{IEEE Transactions on robotics. VOL. 41, 2025},
\maketitle

\begin{abstract}
LiDAR point cloud is essential for autonomous vehicles, but motion distortions from dynamic objects degrade the data quality.
While previous work has considered distortions caused by ego motion, distortions caused by other moving objects remain largely overlooked, leading to errors in object shape and position.
This distortion is particularly pronounced in high-speed environments such as highways and in multi-LiDAR configurations, a common setup for heavy vehicles.
To address this challenge, we introduce HiMo, a pipeline that repurposes scene flow estimation for non-ego motion compensation, correcting the representation of dynamic objects in point clouds.
During the development of HiMo, we observed that existing self-supervised scene flow estimators often produce degenerate or inconsistent estimates under high-speed distortion. 
We further propose SeFlow++, a real-time scene flow estimator that achieves state-of-the-art performance on both scene flow and motion compensation.
Since well-established motion distortion metrics are absent in the literature, we introduce two evaluation metrics: compensation accuracy at a point level and shape similarity of objects.
We validate HiMo through extensive experiments on Argoverse 2, ZOD and a newly collected real-world dataset featuring highway driving and multi-LiDAR-equipped heavy vehicles. Our findings show that HiMo improves the geometric consistency and visual fidelity of dynamic objects in LiDAR point clouds, benefiting downstream tasks such as semantic segmentation and 3D detection.
See \url{https://kin-zhang.github.io/HiMo} for more details.

\end{abstract}

\glsresetall


\section{Introduction}
\IEEEPARstart{L}{ight} \gls{lidar} sensors are an integral part of perception systems for autonomous driving. These sensors provide detailed depth information and point cloud data, complementing other sensing modalities (such as cameras) to offer high-precision 3D scene understanding \cite{pan2024tro,zhong2023icra,zhong2024cvpr,cai2024dmap,zhang2025zero,wang20224d,li2024sscbench}.
However, due to the rotating mechanism of mechanical LiDAR sensors, different parts of the environment are measured at different times.
This introduces motion-induced point cloud distortion in dynamic scenes, making it difficult to obtain accurate environment representations. 
We refer to this as rolling shutter distortion~\cite{yang2023emernerf} to highlight the close relation with the effect seen in the image domain \cite{Cao_2024_CVPR,Qu_2023_ICCV}. 

\begin{figure}[t]
\centering
\includegraphics[trim=200 0 300 0, clip, width=\linewidth]{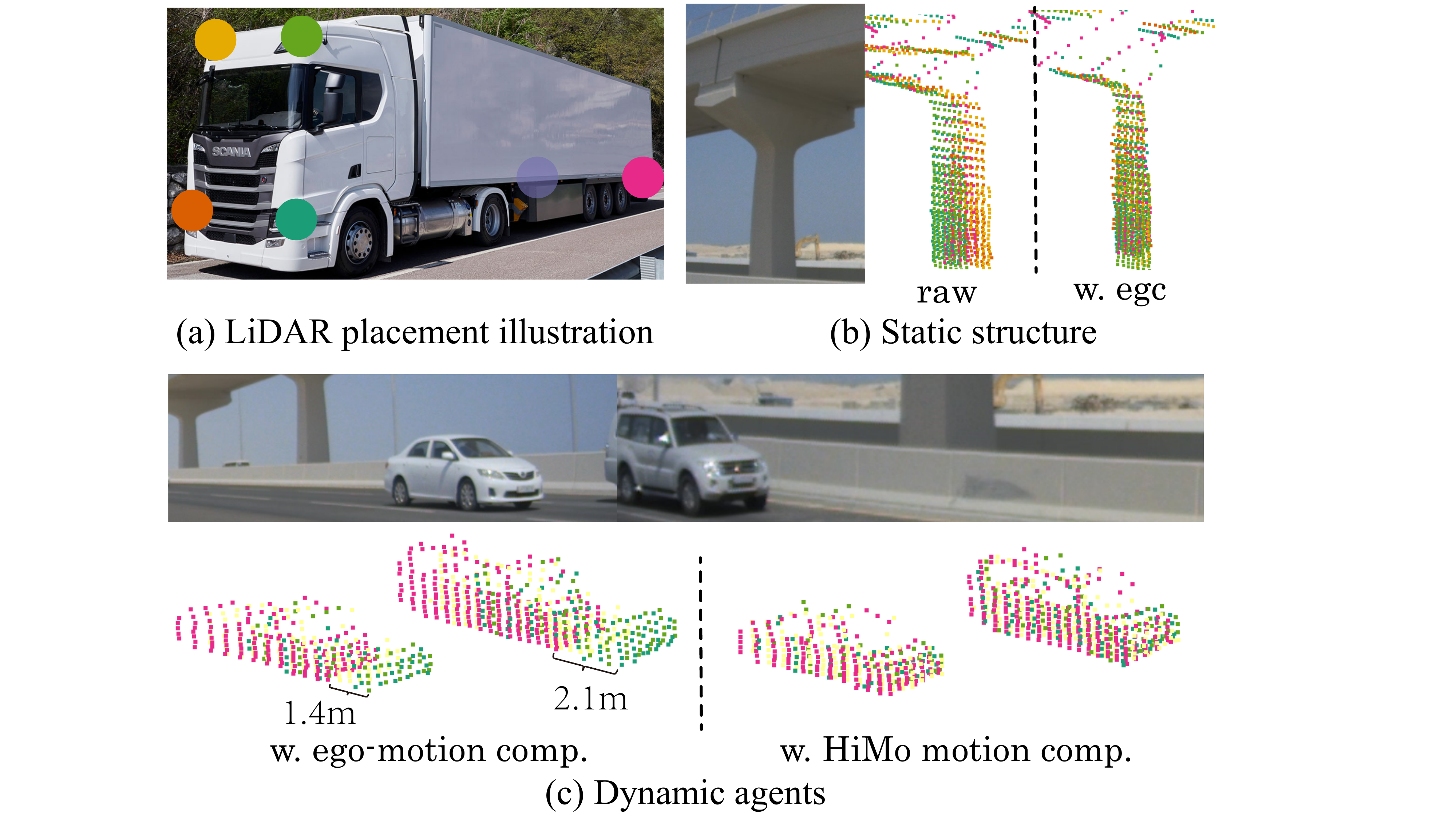}
\caption{
Multi-LiDARs are equipped in our heavy vehicles to avoid self-occlusion. 
(a) shows an example placement with 6 LiDARs. The point colors in (b-c) correspond to the LiDAR from which the points are captured.
(b) illustrates the distortion of static structure due to fast-moving ego vehicle. \textit{Raw} shows the raw data, \textit{w. egc} shows the ego-motion compensation results.
(c) demonstrates distortion caused by motion of other objects, which depends on the velocity of the said objects. 
In such case, ego-motion compensation alone (\textit{w. ego-motion comp.}) is insufficient. 
In comparison, our HiMu pipeline (\textit{w. HiMo motion comp.}) successfully undistorts the point clouds completely, resulting in an accurate representation of the objects.
}
\label{fig:background}
\vspace{-1.0em}
\end{figure}

There are two primary causes for LiDAR rolling shutter distortion: the motion of the ego vehicle and the motion of other agents in the scene. 
In the first case, the movement of the ego-vehicle, combined with the latency caused by the mechanical rotation of the LiDAR, leads to distorted representations of static objects or scenes.
Such ego-motion induced distortion is well-studied in robotics \cite{egomotionmatlab}. 
In practice, localization algorithms \cite{xu2022fast,nguyen2023slict,jiao2021robust} and additional global positioning devices can be employed to accurately correct this distortion (see \cref{fig:background} (b)).

The second source of error -- motion of other agents, on the other hand, is underexplored.
In this case, the motion distortion is object-specific and depends on the relative velocity between the dynamic objects and the ego-vehicle. 
An example of this is shown in \cref{fig:background} (c), where the ego-motion compensated point cloud representation of the gray car is more elongated compared to the white car due to its higher velocity.
This example demonstrates that ego-motion compensation alone cannot effectively address the distortion caused by dynamic objects.
This residual distortion has significant consequences: When sweeps of multiple LiDARs are combined, multiple copies of the same objects may appear in the merged point cloud, leading to incorrect object position or misleading representations that complicate downstream tasks.

\begin{figure}[t]
\centering
\includegraphics[width=\linewidth]{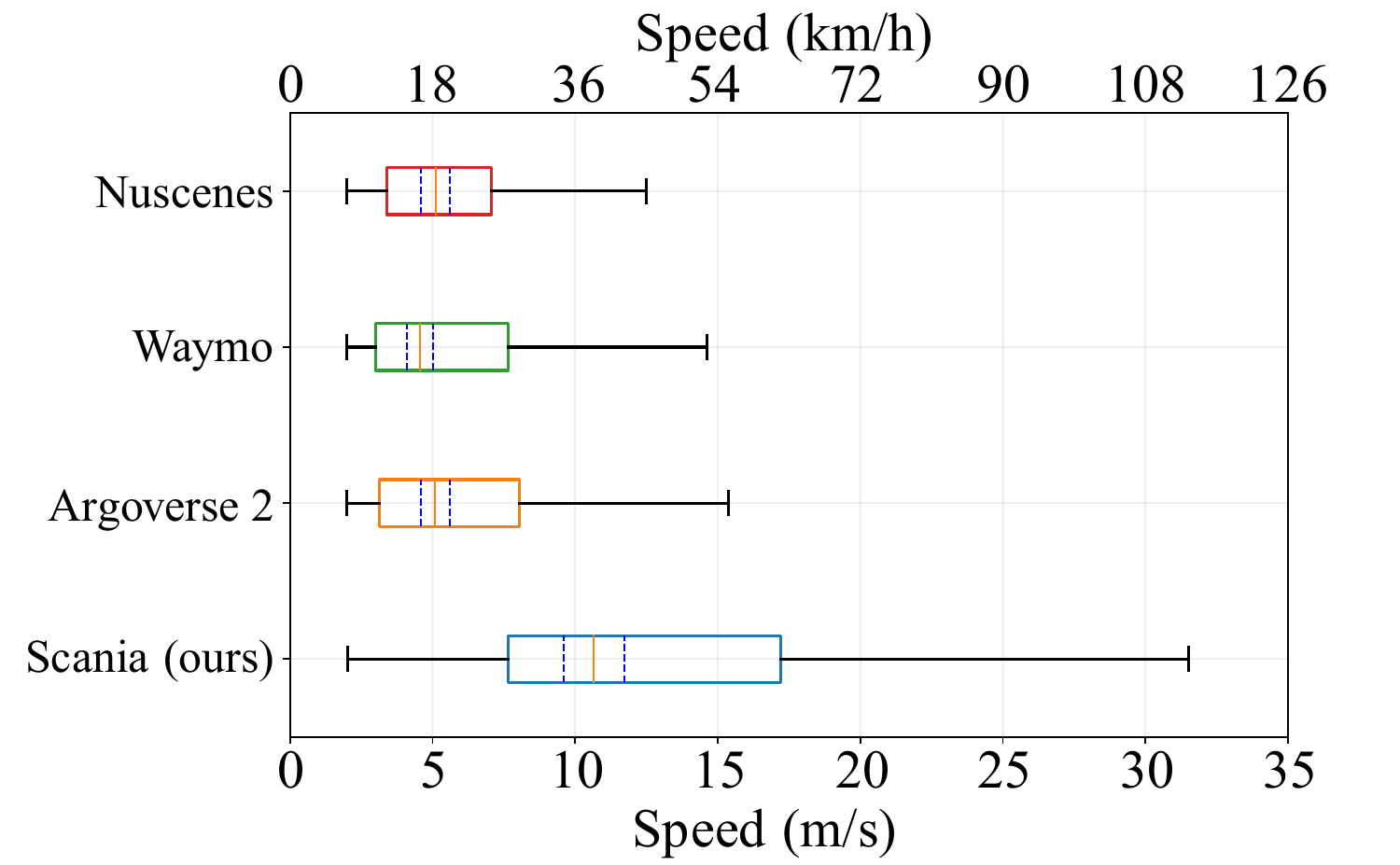}
\caption{
Distribution of object speeds in different datasets. 
Only objects with speeds exceeding 2\si{m/s} are included in the plot.
The orange line shows the median speed and the blue dashed lines indicate the $\pm10\%$ spread.
}
\label{fig:speed_box_plot}
\vspace{-1.0em}
\end{figure}

However, to the best of our knowledge, the non-ego motion distortion caused by dynamic objects has not yet been reported in common public datasets. 
This is likely due to the low object speeds in these datasets. 
Existing open datasets in autonomous driving, such as KITTI \cite{Geiger2013IJRR}, Argoverse \cite{Argoverse,Argoverse2}, Waymo \cite{Waymo} and Nuscenes \cite{nuscenes}, focus on urban environments, where the speeds of most dynamic objects are lower than 40~\si{km/h} (around 11.1 \si{m/s}), as shown in~\cref{fig:speed_box_plot}. 
At such low speed, the distortion exists but is less pronounced (see \cref{sec:distortion} for details).
Nonetheless, such distortions cannot be ignored when bringing autonomous driving solutions to real-world scenarios with large and fast-moving vehicles.

There are multiple research areas that tackle challenges in dynamic point clouds. 
Among them, tasks closely related to ours include moving object segmentation~\cite{sun22mos,cheng2024mf,li2025mos}, which classifies points into dynamic and static categories, and scene flow estimation~\cite{vedula2005three,fastflow3d}, which estimates point-wise motion in successive LiDAR scans.
However, these methods are primarily designed to analyze motion \textit{after} the point cloud has been captured, and do not address \textbf{non-ego motion distortion} that occurs \textit{during} the LiDAR scanning process.
In this paper, we focus on this particular problem and propose \textbf{Hi}gh-speed object \textbf{Mo}tion compensation (HiMo), a pipeline for non-ego motion compensation for point clouds.
An example of HiMo compensated results is shown in~\cref{fig:background}~(c). 
Our primary contributions are as follows:
\begin{itemize} 
\item We provide an in-depth analysis of motion-induced distortions in autonomous driving datasets, and highlight that high-speed objects have larger distortions. 
To support this claim and facilitate further evaluation, we collect \textit{Scania} - a heavy-vehicle highway multi-LiDAR driving dataset, featuring significantly higher average object speed than existing datasets (see \cref{fig:speed_box_plot}).
\item We propose HiMo, the first pipeline for non-ego motion compensation, which leverages self-supervised scene flow estimation to undistort point clouds, significantly improving object representation accuracy. We demonstrate that HiMo compensated point clouds achieve better performance in downstream tasks, including segmentation, 3D detection, and discuss its implications for planning.
\item We additionally develop SeFlow++, a self-supervised scene flow method that enables efficient training via refined auto-labeling and a symmetric Chamfer loss, supporting effective motion compensation for HiMo under limited data and in high-speed conditions. It also achieves state-of-the-art scene flow performance among real-time self-supervised methods.
\item We present two evaluation metrics for non-ego motion compensation and conduct extensive ablation studies with different scene flow estimators to demonstrate the modularity of HiMo and its robustness across datasets.
\end{itemize}

We provide our evaluation data and all codes at \color{blue}\href{https://github.com/KTH-RPL/HiMo}{https://github.com/KTH-RPL/HiMo}\color{black}~to promote the reproducibility and further development of our work.

\section{Related work}

\subsection{Motion Compensation}
As mentioned previously, the motion-induced distortion can be decomposed into two components. 
The first occurs due to the motion of the ego vehicle. 
In robotics, particularly within the field of \gls{slam}, existing methods \cite{xu2022fast,nguyen2023slict} account for this through ego-motion compensation to a specific timestamp. This timestamp is typically chosen to be in the middle of the LiDAR scan, or in the middle of a scan window in case of multiple LiDARs. 
The ego vehicle is typically assumed to be moving at a piece-wise constant velocity, and the coordinate of each point is transformed according to the displacement between the point's timestamp and the motion compensation timestamp. 
This is the baseline motion compensation strategy used in all public autonomous driving datasets~\cite{Geiger2013IJRR,Argoverse2,nuscenes,waymo_perception,alibeigi2023zenseact}.

The second component of this distortion, non-ego motion distortion, occurs due to the motion of other agents in the scene. 
To the best of our knowledge, no existing work addresses the correction of such distortion in raw point cloud data.
While dynamic object segmentation methods~\cite{sun22mos,cheng2024mf,li2025mos} are related to motion understanding, they do not focus on distortion correction.
The recent work SMORE~\cite{chodosh2024simultaneousmapobjectreconstruction} mentions this dynamic rolling shutter distortion in point clouds. 
However, their work focuses on combining multiple frames of distorted data to improve the quality of reconstructed object meshes, without correcting the raw data itself. 
Instead, our work focuses on the distortion correction of \textit{raw data} inside a single LiDAR frame, with many possible downstream applications.
Additionally, the proposed HiMo pipeline is self-supervised, while SMORE requires ground truth tracking labels or a tracking network trained with annotated data.

In summary, our work aims to employ established self-supervised scene flow methods alongside ego-motion compensation to account for all distortions in the raw LiDAR data properly. As such, our method is general and agnostic to the downstream application.

\subsection{Scene Flow Estimation}
Scene flow estimation is the task of describing a 3D motion field between temporally successive point clouds \cite{vedula2005three,jiang20243dsflabelling,khatri2024can,zhang2024gmsf,liu2023difflow3d,Wang_2023_CVPR,yang2023emernerf}. 
Existing works applied to autonomous driving datasets can be categorized into supervised~\cite{fastflow3d,zhang2024deflow,khoche2025ssf,kim2024flow4d} and self-supervised flow estimation~\cite{zhang2024seflow,vedder2024neural,li2021neural,li2023fast,lin2024icp,letitflow,ahuja2024optflow,zhu2023deflow}.

Most supervised networks employ an object detection backbone and connect it to a decoder that generates output flow.
For instance, FastFlow3D \cite{fastflow3d} uses a feedforward architecture based on PointPillars \cite{lang2019pointpillars}, an efficient LiDAR detector architecture, enabling efficient training and inference of flow in the real world. DeFlow \cite{zhang2024deflow} integrates GRU~\cite{cho2014learning,wei2021pv} with iterative refinement in the decoder design for voxel-to-point feature extraction and boosts the performance of flow estimation. Supervised methods require training with detailed ground truth flow labels. Such labels are expensive and time-consuming to gather, limiting the scalability of these methods.

To train models without labeled data or directly optimize at runtime, researchers propose self-supervised pipelines for scene flow estimation \cite{zhang2024seflow,li2021neural,li2023fast,lin2024icp,zeroflow}. 
Neural Scene Flow Prior (NSFP)~\cite{li2021neural} provides high-quality scene flow estimates by optimizing MLP layers at test time to minimize the Chamfer distance and maintain cycle consistency. FastNSF~\cite{li2023fast} leverages the same optimization but achieves significant speedup by computing the Chamfer loss using distance transform \cite{asad2022fastgeodis}.
ICP-Flow \cite{lin2024icp} performs Iterative Closest Point (ICP) between each cluster in two point clouds and trains a feedforward neural network for real-time inference.
SeFlow \cite{zhang2024seflow} integrates dynamic awareness and proposes novel self-supervised loss terms to allow for efficient training using large datasets. 

Existing scene flow estimation methods are not intended to address non-ego motion distortions in raw point cloud data. 
However, they capture point-level velocity information that is valuable for the task. 
Guided by this insight, in this work, we repurpose scene flow estimation as a key component to correct non-ego motion distortions.
In developing HiMo, we observed that existing self-supervised scene flow estimators often produce degenerate or inconsistent estimates under high-speed distortion. 
To mitigate these inconsistencies, we introduce SeFlow++, a scene flow estimator built on SeFlow but with higher training efficiency and better performance, particularly in high-speed scenarios. Specifically, we design a dynamic auto-labelling module (See~\cref{sec:autolabel}) to improve the self-supervised signal at high-speed motion and a symmetric loss computation (See~\cref{sec:selfloss}) with a larger three-frame model backbone to improve training efficiency with limited data.

\begin{figure}[t]
\centering
\begin{subfigure}[t]{0.49\linewidth}
    \centering
    \includegraphics[trim=660 410 750 140, clip, width=\linewidth]{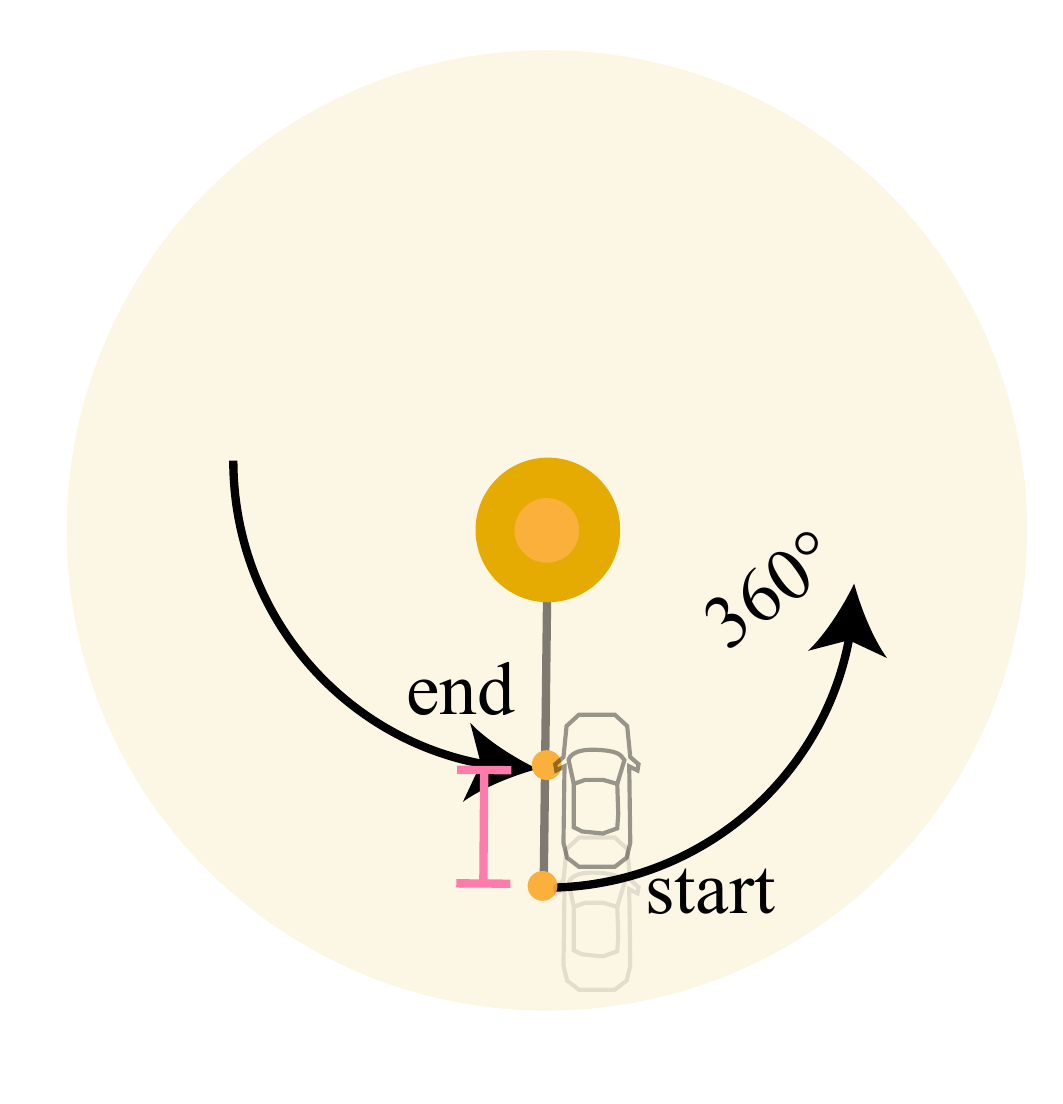}
    \caption{}
    \label{fig:reason_1}
\end{subfigure}
\hfill
\begin{subfigure}[t]{0.49\linewidth}
    \centering
    \includegraphics[trim=1200 410 220 140, clip, width=\linewidth]{img/pdf/reason.pdf}
    \caption{}
    \label{fig:reason_2}
\end{subfigure}
\caption{
Top-view example of LiDAR sweeps showing how distortions are created for vehicles with (a) a single LiDAR and (b) two LiDARs. The small concentric circles of yellow and green are LiDARs. 
Both cases cause a displacement distance for the high-speed object.
The light and dark cars show the vehicle's positions at two timestamps (the beginning and the end of the LidDAR sweep).
(a) One complete single LiDAR scan sweep (small yellow dots are the first and last scan points). This case is only observed when moving objects are at the scan boundaries. 
(b) Two LiDAR scans separated in orientation by 180 degrees. This case is always observed for fast-moving objects. An animation illustrating both cases is included in the supplementary video.}
\label{fig:reason}
\vspace{-0.5em}
\end{figure}

\begin{figure*}[h!]
\centering
\includegraphics[trim=140 280 190 70, clip, width=\linewidth]{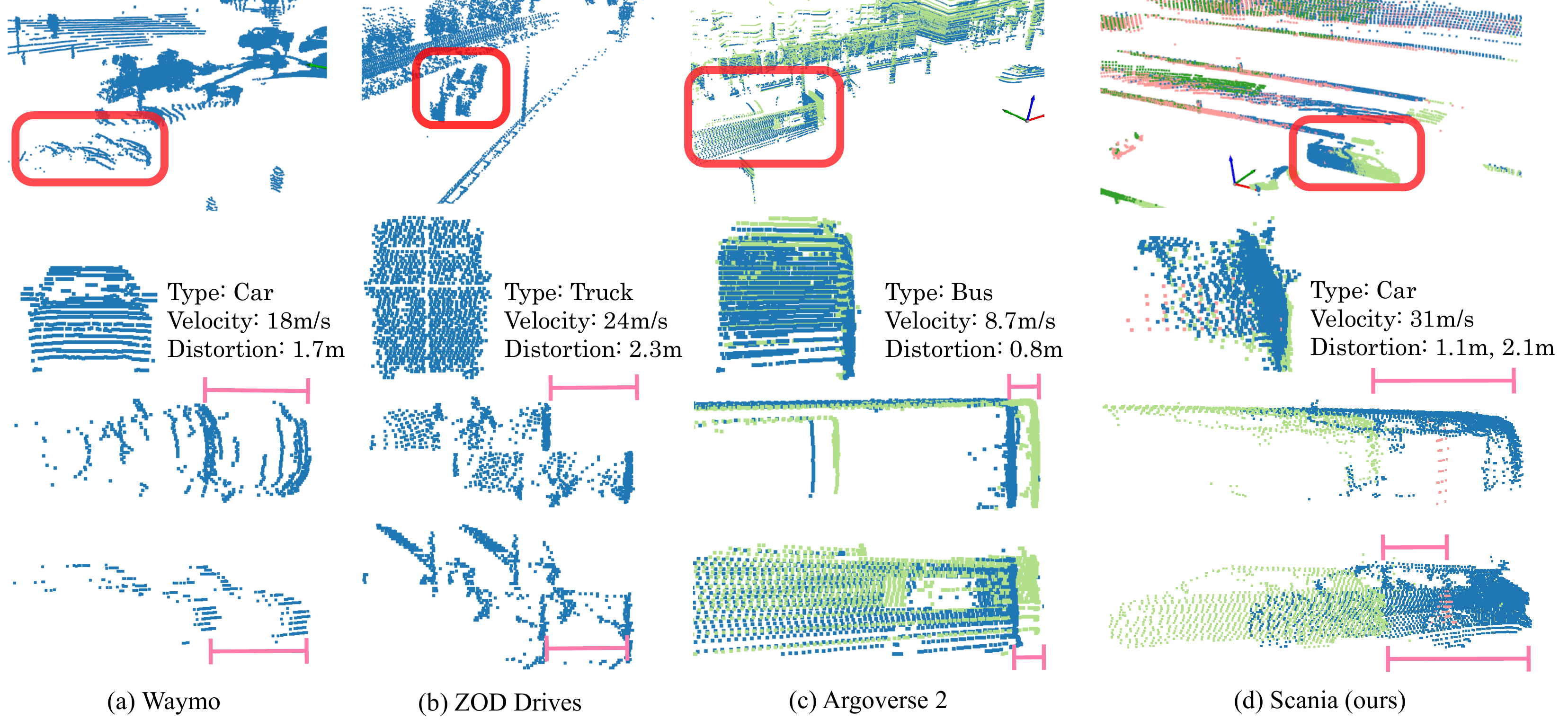}
\caption{
Examples of LiDAR distortion in various datasets after ego-motion compensation \cite{Argoverse2,Waymo,alibeigi2023zenseact}. 
Ground points are removed in visualizations for a clearer view. 
Each column shows an example from a different dataset.
Within each column, the top image shows the full scene, while the three images below show the zoomed-in front, top-down, and side views, respectively.
(a) and (b) showcase the scenario described in \cref{fig:reason_1}, where an object is captured right at the beginning and at the end of the scan in a single LiDAR setup.
(c) and (d) showcase the multi-LiDAR distortion scenario described in \cref{fig:reason_2}.
The different colors in these two subfigures represent data from different LiDARs.
}
\label{fig:data_example}
\vspace{-0.5em}
\end{figure*}

\section{Motion Compensation}
In this section, we begin by discussing the point cloud distortion caused by dynamic objects in autonomous driving.
We then propose our general HiMo pipeline to address this challenge.

\subsection{Non-ego Motion Distortion}
\label{sec:distortion}
Commonly used mechanical LiDAR sensors operate by sweeping laser beams in a horizontal ring pattern. 
This scanning process takes a certain amount of time to complete a full 360-degree sweep. 
If the sensed objects move during the scan, the captured point cloud will be distorted. 
The degree of the distortion is highly correlated with the velocity of the observed object ($v_{\text{object}}$) and sensor frequency ($f_{\text{sensor}}$), with the maximum distorted distance being $v_{\text{object}} / {f_{\text{sensor}}}$. 

\cref{fig:reason} illustrates this distortion in the single-\gls{lidar} and multi-\gls{lidar} scenarios. 
In the single-\gls{lidar} case, this distortion is most visible when the dynamic object is positioned at the edge of the scan (see \cref{fig:reason_1}).
This can lead to severe shape distortion in the resulting point cloud. 
\cref{fig:data_example}(a) and (b) provide examples from Waymo \cite{waymo_perception} and ZOD \cite{alibeigi2023zenseact}, respectively. 

In multi-LiDAR systems, the distortion problem is even more pronounced.
As shown in \cref{fig:reason_2}, multiple LiDARs mounted on the same vehicle capture the same moving object at slightly different times and from different perspectives.
This leads to multiple copies of the same object in the data. 
An example of this can be seen in  \cref{fig:data_example}(c) and (d) in Argoverse 2 \cite{Argoverse2} and our Scania dataset, respectively.
In this figure, each color represents a single-frame data from a separate LiDAR.
The distance between these differently colored copies of the object demonstrates the distortion effect in multi-LiDAR setups.

\subsection{HiMo Pipeline}
\label{sec:himopipeline}
\begin{figure*}[t]
\centering
\includegraphics[width=\linewidth, trim=0 385 0 0, clip]{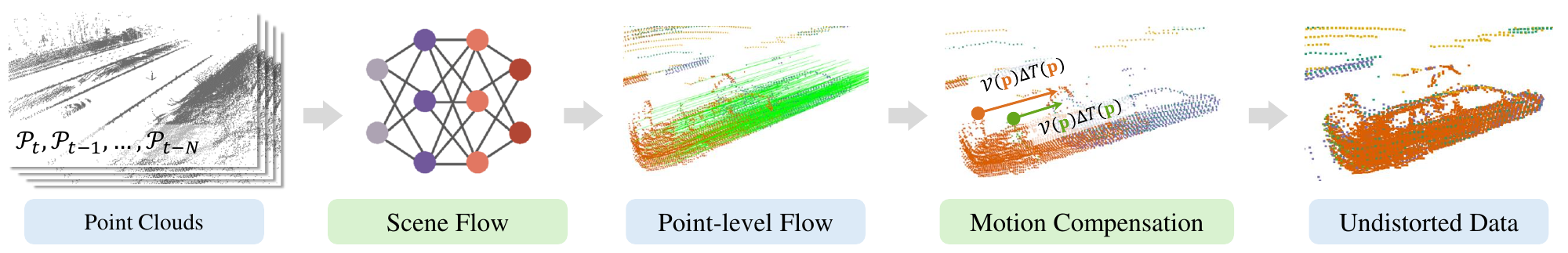}
\caption{
Schematic of the HiMo pipeline. 
Given a sequence of consecutive point cloud frames, a scene flow estimator is employed to calculate the flow of each point. This flow, together with the known LiDAR scan interval and point time difference $\Delta T(\mathbf{p})$, allows us to compute the 3D distortion correction vector. Finally, the undistorted point cloud is computed by combining the correction vector with the raw point cloud.}
\label{fig:pipeline}
\vspace{-1em}
\end{figure*}

In most public datasets, point cloud data is routinely ego-motion compensated \cite{Argoverse2,waymo_perception,alibeigi2023zenseact}.
However, as shown in \cref{fig:background} (c) and \cref{fig:data_example}, this compensation cannot correct the distortions caused by the motion of other agents in the scene. 
To fully compensate for \textbf{all} dynamics in the scene, we propose the following HiMo pipeline. The schematics of the pipeline can be seen in~\cref{fig:pipeline}.

Given the raw input point cloud $\mathcal{P}_{raw}$ from a scene, the goal of the HiMo pipeline is to recover the corresponding point cloud $\mathcal{P}'$ that accurately describes the environment, where all motion-related distortions are corrected.
This $\mathcal{P}'$ can be recovered if we estimate the 3D distortion correction vector for each individual point, denoted as $\mathcal D(\mathbf{p})$.
Using this vector, the estimated undistorted point cloud then can be expressed as $\mathcal{P}' = \mathcal P_{raw} + \mathcal D$.

Note that the 3D distortion correction vector $\mathcal{D}$ can be expressed as follows:
\begin{equation}
    \mathcal{D}(\mathbf{p}) = \mathcal V(\mathbf p) \Delta T(\mathbf{p}),
    \label{eq:repurpose}
\end{equation}
where $\mathcal V(\mathbf p)$ is the velocity of the point, and $\Delta T(\mathbf{p}) \in [0, T_{\text{sensor}} ]$ is the time difference to the timestamp of the last point in this \gls{lidar} scan. $T_\text{sensor}$ denotes the duration of a full LiDAR scan (i.e., a complete 360° sweep), and depends on the scan frequency $f_\text{sensor}$, such that $T_\text{sensor} = 1/f_\text{sensor}$.

Further note that $\mathcal V(\mathbf p)$ can be approximated from the flow of point $p$:
\begin{equation}
    \mathcal V(\mathbf p) = \mathcal{F}(\mathbf{p})/T_{\text{sensor}}   ,
    \label{eq:sf2vel}
\end{equation}
where $\mathcal{F}(\mathbf{p}) = (x,y,z)^T$ is the 3D flow vector of point $\mathbf{p}$ from the current scene to the next. 
In our work, this flow is extracted from the outputs of scene flow estimators. 
Although the goal is to correct distortions within a single LiDAR frame, consecutive frames ($\geq2$) are needed for scene flow estimation.

In summary, HiMo achieves effective motion compensation by repurposing scene flow methods as velocity estimators to compute distortion correction vectors.
As highlighted in~\cref{fig:pipeline}, the pipeline is agnostic to the choice of scene flow estimator, allowing for the adoption of improved scene flow methods as they emerge.

\section{Scene Flow}
Scene flow is the core module in our HiMo motion compensation pipeline. 
To train or optimize a scene flow estimator, we need a supervision signal and a proper loss function. In supervised training, the signal is provided by human-labeled ground truth flow \cite{zhang2024deflow,kim2024flow4d}. 
However, such human annotation is costly and potentially error-prone due to the distortions of high-speed vehicles in raw LiDAR data. 
We therefore focus on self-supervised methods that do not rely on annotations. 
In our experiments, we found that existing self-supervised scene flow methods either require huge computational resources or do not perform well under the conditions of scarce training data and high-speed object distortions.
To adapt to the high-speed regime and reduce the requirement on the amount of training data, we present SeFlow++, a new scene flow method based on the previous state-of-the-art method SeFlow~\cite{zhang2024seflow}. 
Compared to SeFlow, SeFlow++ enhances training efficiency and temporal consistency by combining symmetric loss functions with a larger three-frame model backbone, and additionally provides more robust supervision signals through a refined dynamic auto-labeler.

\subsection{Auto-labeler}
\label{sec:autolabel}
\begin{figure}[t]
\centering
\includegraphics[trim=90 330 770 155, clip, width=\linewidth]{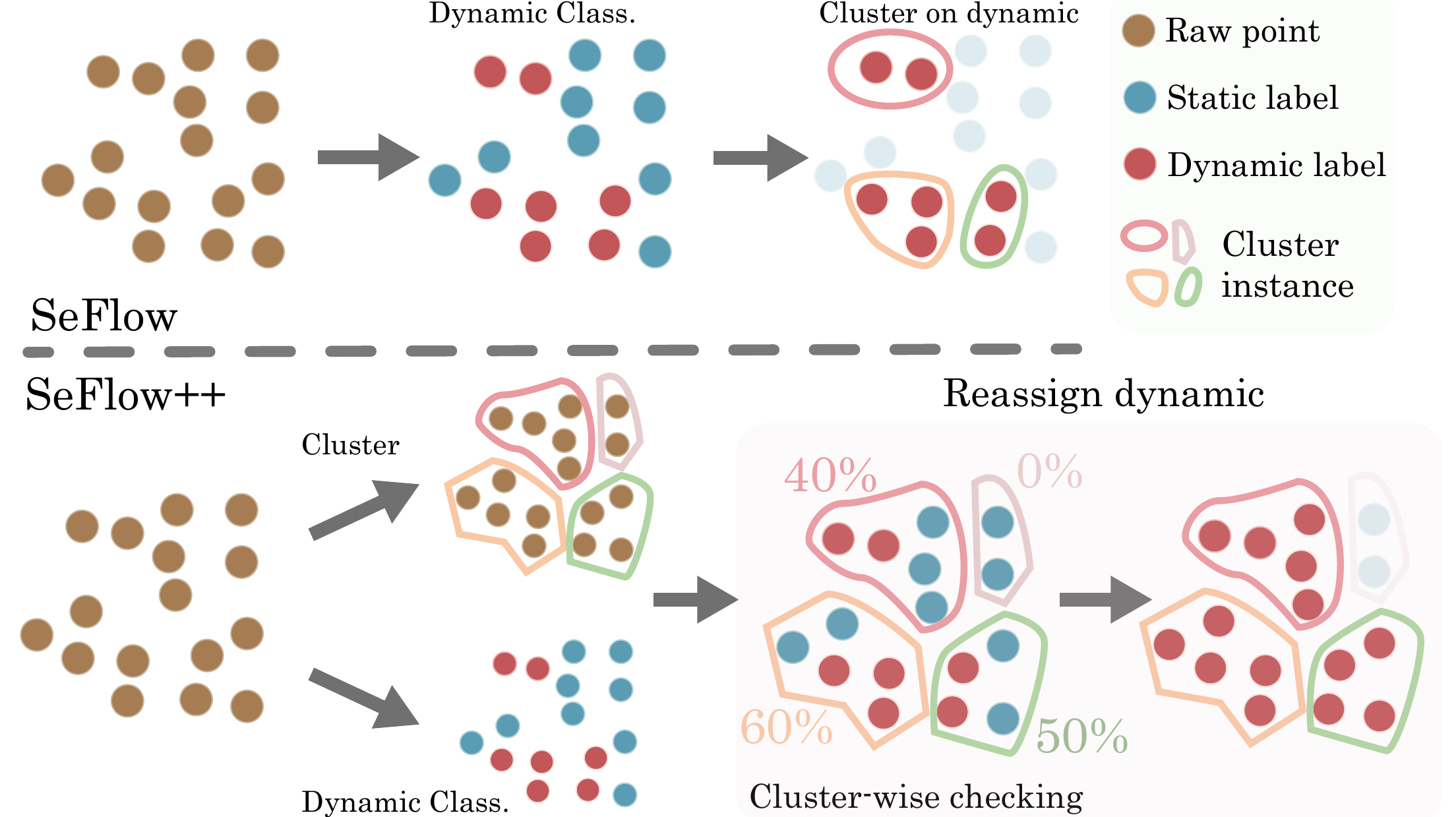}
\caption{
Dynamic auto-labeling process in SeFlow (top) and the improved SeFlow++ (bottom).
In SeFlow, dynamic classification, denoted by \textit{Dynamic Class.}, is first performed by DUFOMap. Then, this classification is used to guide dynamic object clustering. In SeFlow++, on the other hand, we simultaneously perform object clustering and dynamic classification using two independent methods.
The \textit{Cluster-wise checking} step then reassigns labels based on whether the proportion of dynamic points labeled in cluster instances exceeds the thresholds.
}
\label{fig:autolabel}
\vspace{-1.0em}
\end{figure}

To obtain a self-supervision signal, SeFlow~\cite{zhang2024seflow} classifies points into static and dynamic based on DUFOMap~\cite{daniel2024dufomap}. The dynamic points are then grouped into clusters, as shown in \cref{fig:autolabel} (top). 
However, we observe that the scene flow estimator trained using the SeFlow strategy is not always accurate.
This is because errors in the dynamic point classification of DUFOMap get propagated into the subsequent object clustering step, resulting in misclassification of dynamic and static points.
To refine the auto-labeling in SeFlow++, we compute the dynamic clusters by aggregating more information. As shown in \cref{fig:autolabel} (bottom), this process starts with the parallel computation of clusters and point-level dynamic classification:

\subsubsection{Clustering}
We cluster all points in $\mathcal P_t$ using HDBSCAN~\cite{campello2013density} into object instances $\mathcal C = \{c_1, c_2, \cdots, c_n\}$.

\subsubsection{Dynamic classification}:
To improve the robustness of this module, we employ two independent dynamic classification methods: DUFOMap~\cite{daniel2024dufomap} and a threshold-based nearest neighbor method~\cite{najibi2022motion}. 
The key insight of DUFOMap is that points observed inside a region that at any time has been observed as empty must be dynamic.
The empty regions can be inferred using ray-casting. 
The result of DUFOMap for $\mathcal P_t$ is a dynamic point set $\mathcal P_{\text{dufo}}$.
The second method, threshold-based nearest neighbor \cite{najibi2022motion}, requires two consecutive point clouds $\mathcal P_t$ and $\mathcal P_{t+1}$ as input. The dynamic point set is defined by:
\begin{equation}
    \mathcal P_{\text{nnd}} = \{ \mathbf{p} | \mathrm D_{\text{min}}(\mathbf{p},\mathcal P_{t+1}) > \tau_{d}, \mathbf{p} \in \mathcal P_t \},
\end{equation}
where $\mathrm D_{\text{min}}(\mathbf{p},\mathcal P_{t+1})$ represents the distance between point $\mathbf{p}$ and its nearest neighbor in $\mathcal P_{t+1}$, and $\tau_{d}$ is a user defined threshold.

\noindent\textbf{Reassign label}: 
Given the clusters $\mathcal C$ and the two dynamic point classification results $\mathcal P_{\text{dufo}}$ and $\mathcal P_{\text{nnd}}$ on an input point set $\mathcal{P}_t$, the final subset of dynamic points is defined as follows (the subscript $t$ is dropped for readability):
\begin{equation}
\mathcal{P}_d = \bigcup{\left\{\mathbf{p}| f(c) , \mathbf{p} \in \mathcal{P}_{c}, c \in \mathcal{C} \right\}}, 
\label{eq:dynamic}
\end{equation}
where $\mathcal{C}$ is the set of all clusters based on HDBSCAN on $\mathcal P$, $c$ and $\mathbf{p}$ represent an individual cluster and a point in the cluster, respectively. 
The function $f$ integrates the two dynamic labels $\mathcal P_{\text{dufo}}$ and $\mathcal P_{\text{nnd}}$ is defined as follows:
\begin{equation}
f(c) =
\begin{cases}
\tt{true}, & \text{if } \min(r_1,r_2) \geq \tau_1 ~\&~ \max(r_1,r_2) \geq \tau_2 \\
\tt{false}, & \text{otherwise}
\end{cases},
\end{equation}
where $r_1=\frac{|\mathcal{P}_{\text{dufo}}(c)|}{|\mathcal P_c|}$ and $r_2=\frac{|\mathcal{P}_{\text{nnd}}(c)|}{|\mathcal P_c|}$ represent the proportion of dynamic points labeled in the cluster instance $c$ by the two methods respectively. $|\mathcal P|$ denotes the cardinality of (i.e., the number of points in) the point cloud $\mathcal P$. 
Conceptually, this means that we examine each individual object clusters $c$ separately, and assign all points inside this cluster as dynamic only if 
both methods identify a proportion of dynamic points of at least $\tau_1$, and at least one of them provides a stronger signal with a proportion of at least $\tau_2$.
The decision thresholds $\tau_1, \tau_2$ are hyperparameters. 

In summary, this improved auto-labeler incorporates clustering and point-level dynamic classification results to provide more robust and accurate object-level dynamic labels. For each point, these labels provide information on its dynamicness, as well as its object cluster, both of which are utilized in the subsequent self-supervised training.

\subsection{Self-supervised Loss}
\label{sec:selfloss}
The most popular loss in self-supervised scene flow estimation is the Chamfer distance, a standard metric for the shape dissimilarity between two point clouds. The definition is:
\begin{equation}
    \operatorname{CD} ({\mathcal P}_i,\mathcal P_j) = \frac{\sum_{p \in {\mathcal P}_i}\mathrm D_{\min}(p,\mathcal P_j)}{|{\mathcal P}_i|} + \frac{\sum_{p \in \mathcal P_j}\mathrm D_{\min}(p,\mathcal {P}_i)}{|\mathcal P_j|}
\end{equation}

SeFlow~\cite{zhang2024seflow} highlighted the association errors in the Chamfer distance and proposed a more robust loss design with the following four-term loss function:
\begin{equation}
    \mathcal L_{total} = \mathcal L_{\text{cham}} + \mathcal L_{\text{dcham}} + \mathcal L_\text{static} + \mathcal L_{\text{dcls}},
    \label{eq:seflowloss}
\end{equation}
where $\mathcal L_{\text{cham}}$ and $\mathcal L_{\text{dcham}}$ are Chamfer distance loss on all points and only dynamic points, respectively. 
$\mathcal L_{\text{static}}$ and $\mathcal L_{\text{dcls}}$ optimize flow estimation in static points and points inside dynamic object clusters, respectively. 
The three loss terms $\mathcal L_{\text{dcham}}$, $\mathcal L_\text{static}$ and $\mathcal L_{\text{dcls}}$ all require auto-labels from dynamic point classification. 
In SeFlow++, these labels are obtained as per description in \cref{sec:autolabel}.

Although SeFlow achieves state-of-the-art performance in self-supervised scene flow estimation, 
we noticed that it occasionally produces degenerate or inconsistent flow estimation between consecutive frames. 
This inconsistency introduces errors when we apply the estimated flow to correct motion distortions. 
Prior works have introduced cycle consistency to improve scene flow estimation and mitigate degenerate solutions, particularly in sparse point clouds~\cite{mittal2020just, li2021neural}. However, these methods rely on explicit backward flow computation or duplicated networks, which are inefficient. 
To encourage consistent flow estimate with minimal overhead, we propose a symmetric flow loss that can be calculated within a single forward pass.
Note that due to the symmetry around $\mathcal {P}_t$, applying the forward flow with a \textbf{negative sign} ($-$) effectively simulates a backward flow from $\mathcal P_t$ to $\mathcal P_{t-1}$. 
Guided by this insight, we reformulate the first chamfer distance loss in \cref{eq:seflowloss} to a symmetric version as follows:
\begin{equation}
    \mathcal L_{\text{cham}} = \operatorname{CD} (\hat{\mathcal P}_{t+1},\mathcal P_{t+1}) + \operatorname{CD} (\hat{\mathcal P}_{t-1},\mathcal P_{t-1}),
\end{equation}
where $\hat{\mathcal{P}}_{t\pm1} = \mathcal P_t \pm \hat{\mathcal F}_t$.
We also augment the dynamic chamfer distance loss $\mathcal L_{\text{dcham}}$ to a symmetric version, except it only considers points that are classified as dynamic in \cref{eq:dynamic}.

The last two loss terms in~\cref{eq:seflowloss}, the static loss $\mathcal L_{\text{static}}$ and the dynamic cluster loss $\mathcal L_{\text{dcls}}$, are kept unchanged in this work. 
The static loss encourages the model to estimate zero flow for static points:
\begin{equation}
    \mathcal L_\text{static} = \frac{1}{|{\mathcal P}_{s}|}\sum_{\mathbf p \in \mathcal P_{s}}|| \Delta \hat {\mathcal F}(\mathbf p) ||_2^2.
\end{equation}

For the dynamic cluster loss, as in \cite{zhang2024seflow}, the objective is to encourage points within the same dynamic cluster to have similar scene flow. Specifically, we identify the point with the largest distance to its dynamic nearest neighbor in the next frame for each dynamic cluster, which is used to estimate a representative flow for the entire cluster. More specifically, we find the index of the point $p_k$ in cluster $c_i \in \mathcal C_{d}$ with the largest distance to its nearest neighbor point in $\mathcal{P}_{t+1,d}$, i.e.,  
\begin{equation}    
    \kappa_i = \argmax_k \{\mathrm D_{\min}(p_k,\mathcal P_{t+1,d}) | p_k \in \mathcal P_{c_i}\}.
    \label{eq:k}
\end{equation}

We then calculate the upper bound, $\tilde f_{c_i}$ on the flow for cluster $c_i$ as 
\begin{equation}
    \tilde f_{c_i} = p'_{\kappa_i} - p_{\kappa_i},
    \label{eq:f_ci}
\end{equation}
where $p'_{\kappa_i}$ is the nearest neighbor to point $p_{\kappa_i}$ in $\mathcal P_{t+1,d}$.
We use this to drive the estimated flows of cluster $c_i$ towards $\tilde f_{c_i}$ as follows:
\begin{equation}
    \mathcal L_{\text{dcls}}=\frac{1}{|\mathcal{P}_{t,d}|}{\sum_{c_i \in \mathcal {C}_{t,d}} \left( \sum_{p_j \in \mathcal P_{c_i}} || \hat f_{p_j} - \tilde f_{c_i} ||^2_2 \right) }.
\end{equation}

In summary, SeFlow++ improves the SeFlow loss from two aspects. Firstly, it improves the temporal consistency and data efficiency by augmenting $\mathcal L_{\text{cham}}$ and $\mathcal L_{\text{dcham}}$ with symmetric flow computation. Secondly, it refines the dynamic auto-labeling using the procedure described in \cref{sec:autolabel}.

\section{Experiments Setup}
\subsection{Dataset}
Experiments are conducted mainly on two large-scale autonomous driving datasets: our highway dataset (Scania) and Argoverse 2~\cite{Argoverse2}. 
Additionally, qualitative results from the Zenseact Open Dataset~\cite{alibeigi2023zenseact} (ZOD) are presented.
For all datasets, ground removal is performed using line-fit ground segmentation~\cite{himmelsbach2010fast}.

\begin{figure}[t]
\centering
\begin{subfigure}{0.8\linewidth}
    \centering
    \includegraphics[trim=100 0 0 0, clip, width=\textwidth]{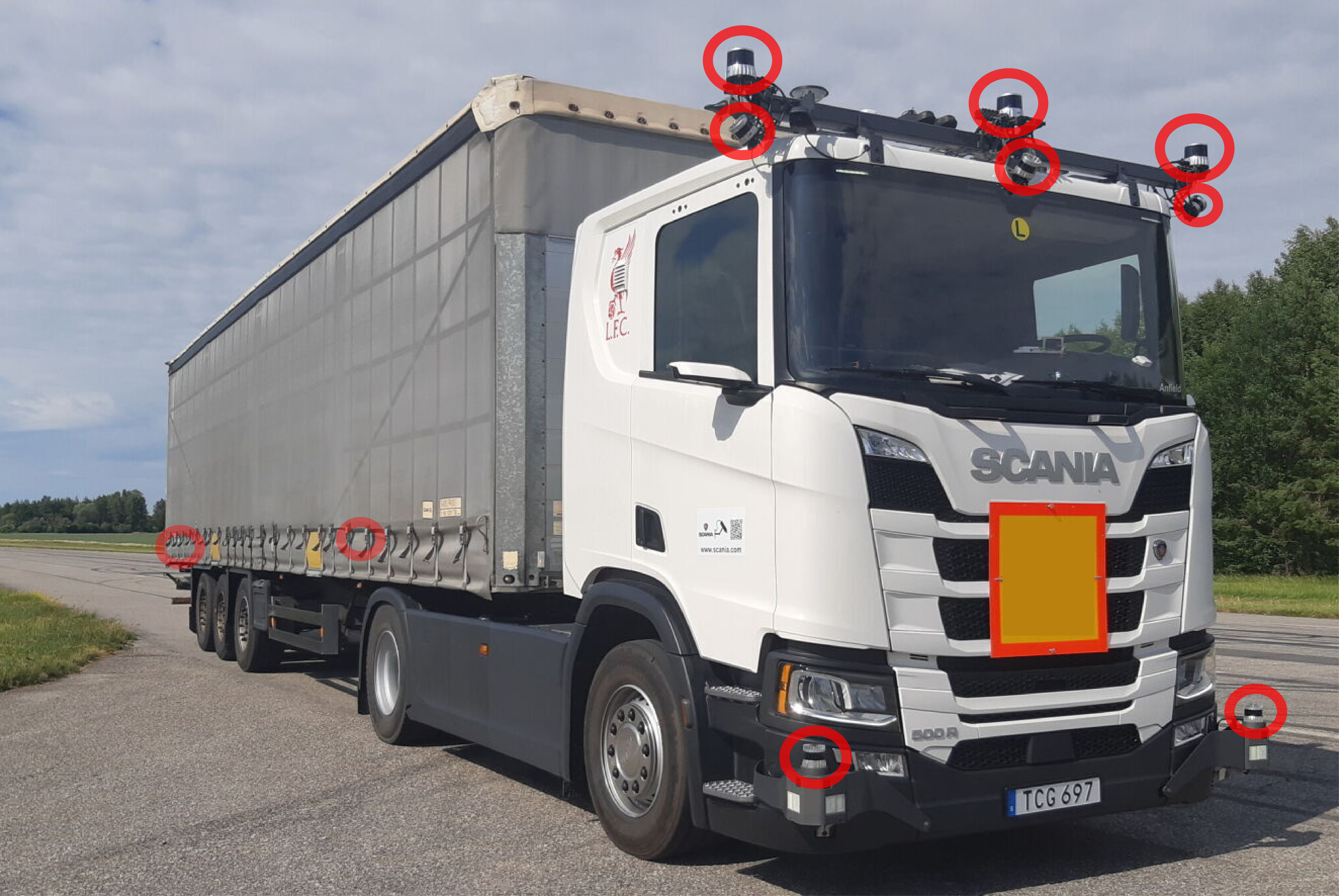}
    \caption{Scania Truck}
\end{subfigure}
\begin{subfigure}{0.8\linewidth}
    \centering
    \includegraphics[width=\textwidth]{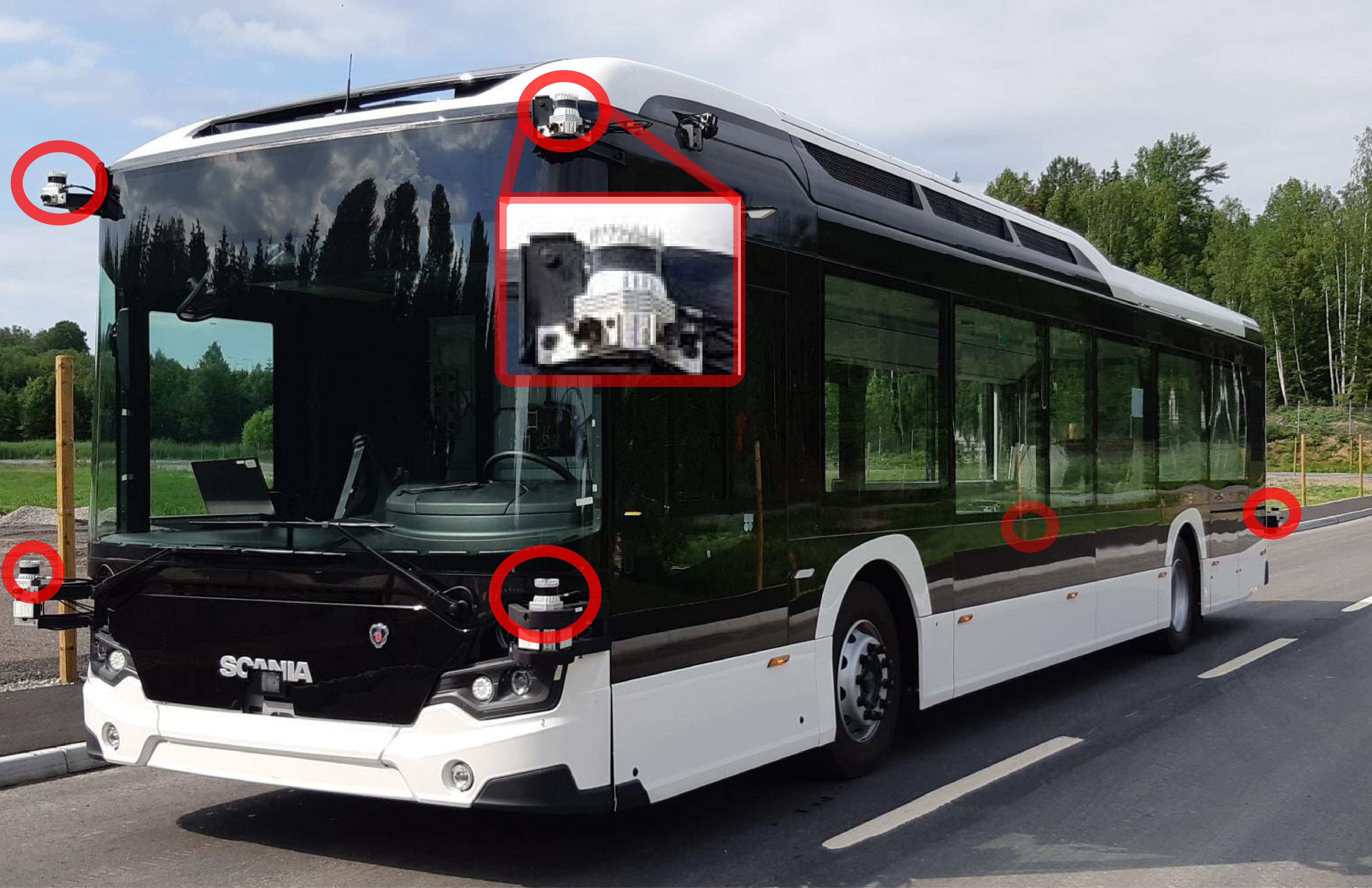}
    \caption{Scania Bus}
\end{subfigure}
\caption{
Our Scania dataset contains data collected from two different multi-LiDAR setups. 
(a) A truck with six LiDARs on the top of the truck and two on the front bottom. There are two at the back of the truck that are not visible in the image. 
(b) A bus with four in the front of the vehicle and two in the back of the bus. 
}
\label{fig:scania}
\vspace{-1em}
\end{figure}

\noindent\textbf{Scania} 
Our dataset consists of around 500 sequences, 10 to 15 seconds per sequence, captured in and around downtown Södertalje, Sweden. 
The total training dataset consists of 48,936 frames from 500 sequences. 
The validation set consists of 100 frames from 10 sequences containing high-speed scenarios. 
Our platforms, as shown in \cref{fig:scania}, consist of buses and trucks equipped with multiple 32-channel \glspl{lidar}. The frequency of each \gls{lidar} is around 10\si{Hz}. Points from multiple \glspl{lidar} within a fixed time interval ($T_{sensor}$) are combined into a single point cloud.

\noindent\textbf{Argoverse 2} Its \textit{Sensor} dataset encompasses 700 training (110,071 frames) and 150 validation sequences. A passenger car equipped with two roof-mounted VLP-32C lidar sensors running at 10 Hz is used for data collection. 
Each scene is approximately 15-20 seconds long, with complete annotations for evaluation. 
As shown in \cref{fig:speed_box_plot}, the average object speed in Argoverse 2 is relatively slow. 
To focus the evaluation on high-speed object motion compensation, the evaluation sequence is selected based on whether it includes at least three fast-moving annotated objects.

More interactive visualization results on Argoverse 2 \cite{Argoverse2} and ZOD \cite{alibeigi2023zenseact} are available on the project page\footnote{\url{https://kin-zhang.github.io/HiMo}}.

\subsection{Annotated Ground truth}
\label{sec:main_anno_gt}
To evaluate the non-ego motion compensation performance, we manually annotated validation data and created undistorted ground truth data through a multi-step process. 
A flowchart of the annotation and refinement pipeline is provided in~\cref{sec:anno_gt}.

Recall from \cref{eq:repurpose} that the 3D distortion correction vector $\mathcal{D}$ can be computed using the velocity $\mathcal{V}$ and the time difference $\Delta T$. 
We can therefore compute the ground truth using the annotated point velocities, which we obtain through object tracking annotations. 
For Argoverse 2, the official sensor annotations already contain manually labelled object bounding boxes and their tracking IDs. 
For Scania, we obtained these by manual labeling of bounding boxes using the protocol in~\cite{khoche2024addressing}. 
However, since we chose to evaluate high-speed scenarios, the distortions in the raw data are significant enough to cause errors in the manual object labels. 
To mitigate these errors, we enlarged the bounding boxes in the direction of motion to include all object copies. 
These refined object tracking labels, combined with the known scanning timestamps and sensor frequency, were used to generate the motion compensation ground truth. 
To ensure geometric consistency, we manually reviewed the ground truth in each frame before including it in the validation set.
\newcommand{\blue}[1]{$_{\color{TableBlue}\downarrow #1}$}
\begin{table*}[t!]
\centering
\def\arraystretch{1.3}
\caption{Object motion compensation result comparisons using different scene flow methods in our HiMo pipeline on the Scania validation set. 
The first row reports the errors for raw data with ego-motion compensation only, and the rest is our HiMo pipeline motion compensation with different scene flow estimators as an ablation study.
Upper groups are supervised methods with scene flow networks trained on Argoverse 2 \cite{Argoverse2} and inference directly on the Scania dataset. 
Lower groups are self-supervised methods. $\pm$ means the standard deviation in the evaluation data (100 frames).
All methods in the HiMo pipeline achieve better accuracy performance and object shape similarity compared to raw data. 
Our proposed SeFlow++ achieves state-of-the-art compensated performance in shape description for both car and other vehicle object types. The \blue{blue} value represents the error percentage decrease relative to the distorted raw data.
}
\begin{tabular}{clc|ccc|ccc} 
\toprule
\multicolumn{2}{c}{\multirow{2}{*}{Methods}} & \multirow{2}{*}{Reference} & \multicolumn{3}{c|}{Chamfer Distance Error (CDE) ↓}                      & \multicolumn{3}{c}{Mean Point Error (MPE) ↓}                        \\
\multicolumn{2}{c}{}                        &                            & Total          & CAR                  & OTHERS      & Total          & CAR                  & OTHERS       \\ 
\hline
\rowcolor[rgb]{0.91,0.91,0.91} \multicolumn{2}{c}{Ego-motion Compensation}         & -                          & \multicolumn{0}{l}{0.284}          & 0.257 $\pm$ 0.13          & 0.310 $\pm$ 0.11          & \multicolumn{0}{l}{0.935}          & 0.913 $\pm$ 0.16          & 0.957 $\pm$ 0.07           \\ 
\hline
\multirow{7}{*}{\begin{tabular}[c]{@{}c@{}}HiMo\\(Ours)\end{tabular}} & FastFlow3D~\cite{fastflow3d}                & RAL'22                    & 0.144 \blue{49\%}         & 0.121 $\pm$ 0.04          & 0.168 $\pm$ 0.04          & 0.546 \blue{42\%}          & 0.378 $\pm$ 0.16          & 0.714 $\pm$ 0.17           \\
 & DeFlow~\cite{zhang2024deflow}               & ICRA'24                   & 0.088 \blue{69\%} & \underline{0.057} $\pm$ 0.02  & 0.118 $\pm$ 0.05          & 0.315 \blue{66\%}  & \textbf{0.139} $\pm$ 0.08 & 0.491 $\pm$ 0.19           \\ 
\cline{2-9}
 & NSFP~\cite{li2021neural}                    & NeurIPS'21                & \underline{0.073} \blue{74\%}  & 0.064 $\pm$ 0.07          & 0.083 $\pm$ 0.04          & \textbf{0.255} \blue{73\%} & 0.188 $\pm$ 0.16          & \underline{0.323} $\pm$ 0.15   \\
 & FastNSF~\cite{li2023fast}                   & ICCV'23                   & 0.078 \blue{72\%} & 0.074 $\pm$ 0.05          & \underline{0.081} $\pm$ 0.05  & 0.279  \blue{70\%}        & 0.251 $\pm$ 0.16          & \textbf{0.308} $\pm$ 0.14  \\
 & ICP-Flow~\cite{lin2024icp}                  & CVPR'24                   & 0.183 \blue{36\%} & 0.203 $\pm$ 0.13          & 0.163 $\pm$ 0.05          & 0.695   \blue{26\%}       & 0.698 $\pm$ 0.21          & 0.692 $\pm$ 0.18           \\
 & SeFlow~\cite{zhang2024seflow}               & ECCV'24                   & 0.096  \blue{66\%} & 0.094 $\pm$ 0.04          & 0.098 $\pm$ 0.01          & 0.452   \blue{52\%}       & 0.444 $\pm$ 0.17          & 0.461 $\pm$ 0.18           \\
 & SeFlow++ (Ours)                             & -                          & \textbf{0.054} \blue{81\%} & \textbf{0.050} $\pm$ 0.03 & \textbf{0.059} $\pm$ 0.02 & \underline{0.267}  \blue{72\%} & 0.179 $\pm$ 0.10          & 0.356 $\pm$ 0.18           \\
\bottomrule
\end{tabular}
\label{tab:main_res}
\vspace{-1.0em} %
\end{table*}

\subsection{Evaluation Metrics}
Due to the lack of well-established motion distortion metrics in the literature, we present two metrics: one captures shape similarity inspired by 3D reconstruction, and the other measures point-level accuracy similar to end point error in scene flow.

Shape similarity measures the correctness of shape descriptions. 
We quantify this similarity using the Chamfer distance error (CDE) defined as:
\begin{equation}
    \text{CDE} = \frac{1}{ |\mathcal{C}_{gt}|}\sum_{c_i \in \mathcal{C}_{gt}} \left(\frac{|\mathcal{P}_{c_i}|}{|\mathcal{P}_{\mathcal C_{gt}}|} \text{CD}(\mathcal {P'}_{est,c_i},\mathcal {P'}_{gt, c_i})\right),
    \label{eq:opti_cds}
\end{equation}
where $|\cdot|$ denotes cardinality of a set, $\mathcal{C}_{gt}$ denotes the ground truth clusters in the frame, $\mathcal{P}_{c_i}$ and $\mathcal{P}_{\mathcal C_{gt}}$ denote the set of points in cluster $c_i$ and in all ground truth clusters, $\mathcal {P'}_{est, c_i}$ and $\mathcal {P'}_{gt, c_i}$ denote the point set $c_i$ compensated using the estimated motion and ground truth, respectively.
The smaller the CDE, the greater the shape similarity between the estimated instance shape and the ground truth. 

While we believe that shape similarity best measures a method's ability to undistort moving objects, its reliance on Chamfer distance implies that it is based on nearest neighbor matching. Such matching results do not guarantee correct association at point level.
Therefore, we also compute the point-level accuracy that can be represented as the mean point error (MPE),
\begin{equation}
    \text{MPE} = \frac{\sum_{c_i \in \mathcal{C}_{gt}}\left(\sum_{\mathbf{p} \in \mathcal{P}_{c_i}} \left\| \mathbf{p}_{est} - \mathbf{p}_{gt} \right\|_2\right)}{ |\mathcal{C}_{gt}||\mathcal{P}_{\mathcal C_{gt}}|}
    \label{eq:opti_mpe}
\end{equation}
where $\mathbf{p}_{est}$ and $\mathbf{p}_{gt}$ denote the estimated and corresponding ground truth point, respectively. The rest of the notations are the same as per CDE.

Note that these two metrics have different focuses: CDE captures the error at an object level, and quantifies the correctness of the undistorted object shape, whilst MPE metric captures the errors at a point level. 

Since the motion distortion in high-speed objects is more pronounced (see \cref{sec:distortion} and \cref{fig:data_example}), we separately report metrics for two types of vehicle categories.
In all result tables, CAR means regular and passenger vehicles, and OTHER VEHICLES (OTHERS) include trucks, long buses, heavy vehicles, vehicles with trailers, etc.

\subsection{Evaluated Methods}
In this work, we address the problem of non-ego motion distortion compensation in LiDAR point clouds, which has not been tackled by prior methods. The only existing baseline for motion compensation is ego-motion compensation, where the point cloud data is adjusted solely for the motion of the ego vehicle. 
In this baseline, the distortion correction vector $\mathcal D$ is the velocity of the ego vehicle at the current timestamp, scaled by $\Delta T(p)$ for each point.

To address non-ego motion distortion compensation, we propose the HiMo pipeline (see \cref{fig:pipeline}), which leverages scene flow estimation to perform motion compensation. 
Given consecutive frames of point cloud data, HiMo employs any scene flow estimators to estimate point velocities and then transforms these velocities into distortion distances. These distortion distances then can be used to undistort all motions in the raw point cloud.

As mentioned in \cref{sec:himopipeline}, the HiMo pipeline is agnostic to the choice of scene flow estimators. 
To evaluate the performance and highlight the flexibility of HiMo, we incorporate different state-of-the-art scene flow estimators into the pipeline.
The evaluated scene flow estimators\footnote{The scene flow performance of these scene flow methods is provided in~\cref{sec:sf_res}.} are outlined below:
\begin{enumerate}
\item FastFlow3D \cite{fastflow3d}: A supervised model trained on the Argoverse 2 sensor dataset.
\item DeFlow \cite{zhang2024deflow}: A supervised model featuring a voxel-to-point flow decoder with refinement, also trained on the Argoverse 2 sensor dataset.
\item NSFP \cite{li2021neural}: A runtime optimization method that uses Chamfer distance between $\mathcal P_{t}+\hat{\mathcal{F}_{t}}$ and $\mathcal P_{t+1}$. It needs thousands of iterations to optimize a simple neural network to output the flow of each new frame.
\item FastNSF \cite{li2023fast}: A runtime speedup version of NSFP that uses a distance transform to calculate the nearest neighbor error.
\item ICP-Flow \cite{lin2024icp}: This approach first processes point clouds with a clustering algorithm. Then it employs the conventional Iterative Closest Point (ICP) algorithm that aligns the clusters over time and outputs the corresponding rigid transformations between frames.
\item SeFlow \cite{zhang2024seflow}: This method integrates efficient dynamic classification into a learning-based scene flow pipeline and designs three novel losses to achieve self-supervised flow training as described in \cref{eq:seflowloss}.
\item SeFlow++ (Ours): Proposed in this paper, it improves SeFlow with refined dynamic auto-labeling 
(see \cref{sec:autolabel}). It leverages a three-frame DeFlow backbone to implement a symmetric self-supervised loss (see \cref{sec:selfloss}), enhancing flow consistency and training efficiency.
\end{enumerate}

\begin{table*}[t!]
\centering
\def\arraystretch{1.3}
\caption{Object motion compensation result comparisons using different scene flow methods in our HiMo pipeline on Argoverse 2. 
The first row reports the errors for raw data with ego-motion compensation only, 
and the rest also includes compensation for non ego-motion distortion using HiMo.
Self-supervised methods are listed without human-labeled data needed to undistort raw data. $\pm$ means the standard deviation in the evaluation data (100 frames).
All methods in the HiMo pipeline achieve better accuracy performance and object shape similarity compared to raw data. 
The \blue{blue} value represents the error percentage decrease relative to the distorted raw data.
}
\begin{tabular}{clc|ccc|ccc} 
\toprule
\multicolumn{2}{c}{\multirow{2}{*}{Methods}} & \multirow{2}{*}{Reference} & \multicolumn{3}{c|}{Chamfer Distance Error (CDE) ↓}                      & \multicolumn{3}{c}{Mean Point Error (MPE) ↓}                        \\
\multicolumn{2}{c}{}                        &                            & Total          & CAR                  & OTHERS      & Total          & CAR                  & OTHERS       \\ 
\hline
\rowcolor[rgb]{0.91,0.91,0.91} \multicolumn{2}{c}{Ego-motion Compensation}         & -                          & \multicolumn{0}{l}{0.180}          & 0.176 $\pm$ 0.02          & 0.185 $\pm$ 0.01          & \multicolumn{0}{l}{0.619}          & 0.585 $\pm$ 0.13          & 0.654 $\pm$ 0.03           \\ 
\hline
\multirow{5}{*}{\begin{tabular}[c]{@{}c@{}}HiMo\\(Ours)\end{tabular}} & NSFP~\cite{li2021neural}                   & NeurIPS'21                & 0.052 \blue{71\%}         & 0.073 $\pm$ 0.03          & 0.032 $\pm$ 0.01          & 0.144  \blue{77\%}        & 0.209 $\pm$ 0.12          & \underline{0.079} $\pm$ 0.02   \\
 & FastNSF~\cite{li2023fast}                  & ICCV'23                   & 0.079 \blue{56\%}         & 0.103 $\pm$ 0.03          & 0.054 $\pm$ 0.00          & 0.260 \blue{58\%}         & 0.331 $\pm$ 0.14          & 0.190 $\pm$ 0.00           \\
 & ICP-Flow~\cite{lin2024icp}                 & CVPR'24                   & 0.053 \blue{71\%}          & 0.060 $\pm$ 0.03          & 0.046 $\pm$ 0.00          & 0.135 \blue{78\%}         & 0.168 $\pm$ 0.13          & 0.101 $\pm$ 0.00           \\
 & SeFlow~\cite{zhang2024seflow}              & ECCV'24                   & \underline{0.040} \blue{78\%}  & \underline{0.041} $\pm$ 0.01  & \textbf{0.039} $\pm$ 0.00 & \underline{0.073} \blue{88\%}  & \underline{0.059} $\pm$ 0.02  & 0.088 $\pm$ 0.01           \\
 & SeFlow++ (Ours)                            & \multicolumn{1}{l|}{}      & \textbf{0.038} \blue{79\%} & \textbf{0.037} $\pm$ 0.01 & \underline{0.040} $\pm$ 0.00  & \textbf{0.067} \blue{89\%} & \textbf{0.058} $\pm$ 0.02 & \textbf{0.077} $\pm$ 0.00  \\
\bottomrule
\end{tabular}
\label{tab:main_av2}
\end{table*}

\begin{figure*}[t]
\centering
\begin{subfigure}[t]{0.45\linewidth}
    \centering
    \includegraphics[width=\textwidth]{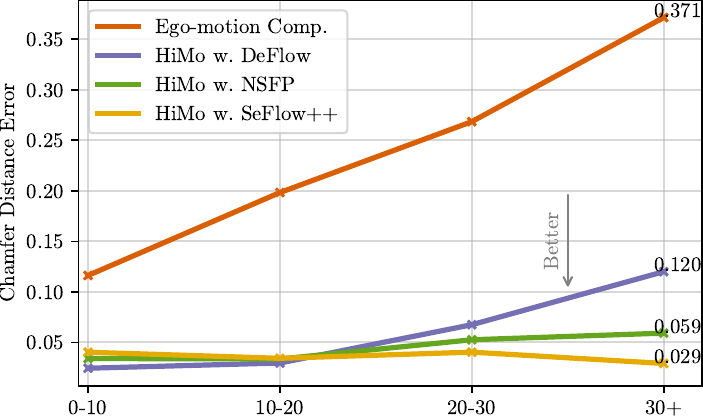}
    \caption{Chamfer distance error v.s. object velocity (\si{m/s})}
    \label{fig:car_dis}
\end{subfigure}
\hspace{10pt}
\begin{subfigure}[t]{0.45\linewidth}
    \centering
    \includegraphics[width=\textwidth]{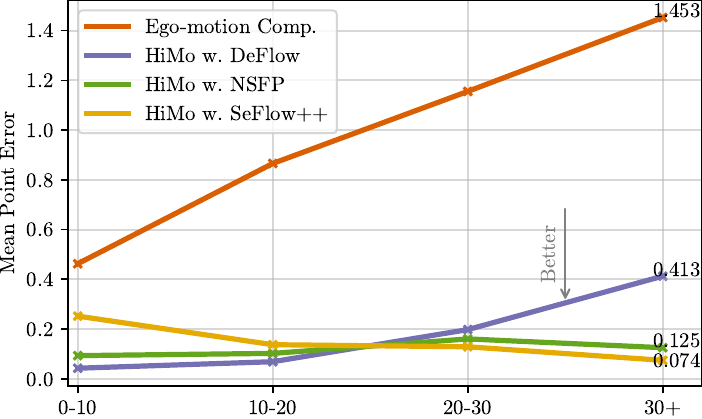}
    \caption{Mean point error v.s. the object velocity (\si{m/s})}
    \label{fig:car_vel}
\end{subfigure}\hspace{5pt}%
\caption{Error distribution concerning object velocity (\si{m/s}) of CAR category for the three best performing flow estimators in our HiMo pipeline on Scania data. The approximately linear relationship between velocity and error is clear in the ego-motion compensated data.}
\label{fig:err_object}
\vspace{-1em}
\end{figure*}

All code to reproduce results and run the HiMo pipeline can be found in~\url{https://github.com/KTH-RPL/HiMo}. 
The main training hyperparameters are listed here for Scania highway datasets: learning rate ($2e^{-4}$) with Adam optimizer \cite{kingma2014adam}, batch size ($8$), the total training epoch ($9$). 
More configuration details can be found in the code. 
All runtime experiments are executed on a desktop powered by an Intel Core i9-12900KF CPU and equipped with a GeForce RTX 3090 GPU. 

\section{Results and Discussion}
\subsection{Quantitative Results}
The comparative analysis of different scene flow methods using the HiMo pipeline on the Scania dataset is detailed in \cref{tab:main_res}. 
The baseline results from \textit{Ego-motion Compensation} are shown on the first row, with large values on both CDE and MPE.
Perfect undistortion would result in both metrics being zero. 
As shown in the rest of \cref{tab:main_res}, compared to this baseline, the HiMo pipeline reduces distortion errors in both CDE and MPE regardless of the scene flow estimator it is coupled with.
This demonstrates the effectiveness of our HiMo pipeline in motion compensation, with up to 81\% in shape improvement. 
However, when combined with different scene flow estimators, the performance of the HiMo pipeline differs.
Despite not having seen Scania data before, DeFlow performs well on the CAR category according to both CDE and MPE. 
This is because the car-type objects in Argoverse 2 and Scania data are similar in shape. 
Comparatively, DeFlow does not perform as well on the OTHERS category. This is because the OTHERS category contains objects with high motion speed and long vehicle sizes, with heavier motion-related distortions in raw data. Such distortions cause difficulty in transferring the knowledge from the previous dataset. 
Compared to supervised methods, most self-supervised methods achieve both lower CDE and MPE on the OTHERS category. 
The only exception is ICP-Flow, which performs on par or worse than the supervised methods on both CAR and OTHERS categories.
This is because ICP-Flow employs many heuristics in its optimization and ICP matching procedure. Hence, extensive parameter tweaking is needed for different data and scenarios.
As shown in \cref{tab:main_av2}, ICP-Flow performs much better on Argoverse 2, the dataset it is optimized for.
The best performance in shape similarity, quantified by the lowest CDE, is achieved by the HiMo pipeline with our proposed SeFlow++. This setup also shows competitive performance on MPE.

To highlight the effectiveness of HiMo, we also provide a quantitative evaluation of the pipeline on the public Argoverse 2 dataset in \cref{tab:main_av2}. 
Comparing the baseline \textit{ego-motion compensation} results on Scania in \cref{tab:main_res} and Argoverse 2 in \cref{tab:main_av2}, we can notice that Argoverse 2's baseline has both lower CDE and MPE. 
This is caused by the lower object speeds as shown in \cref{fig:speed_box_plot}.
As shown in \cref{tab:main_av2}, the HiMo pipeline achieves at least $50\%$ error reduction when combined with any of the five tested self-supervised scene flow estimators.

\begin{figure*}[h]
    \centering
    \includegraphics[width=0.99\linewidth]{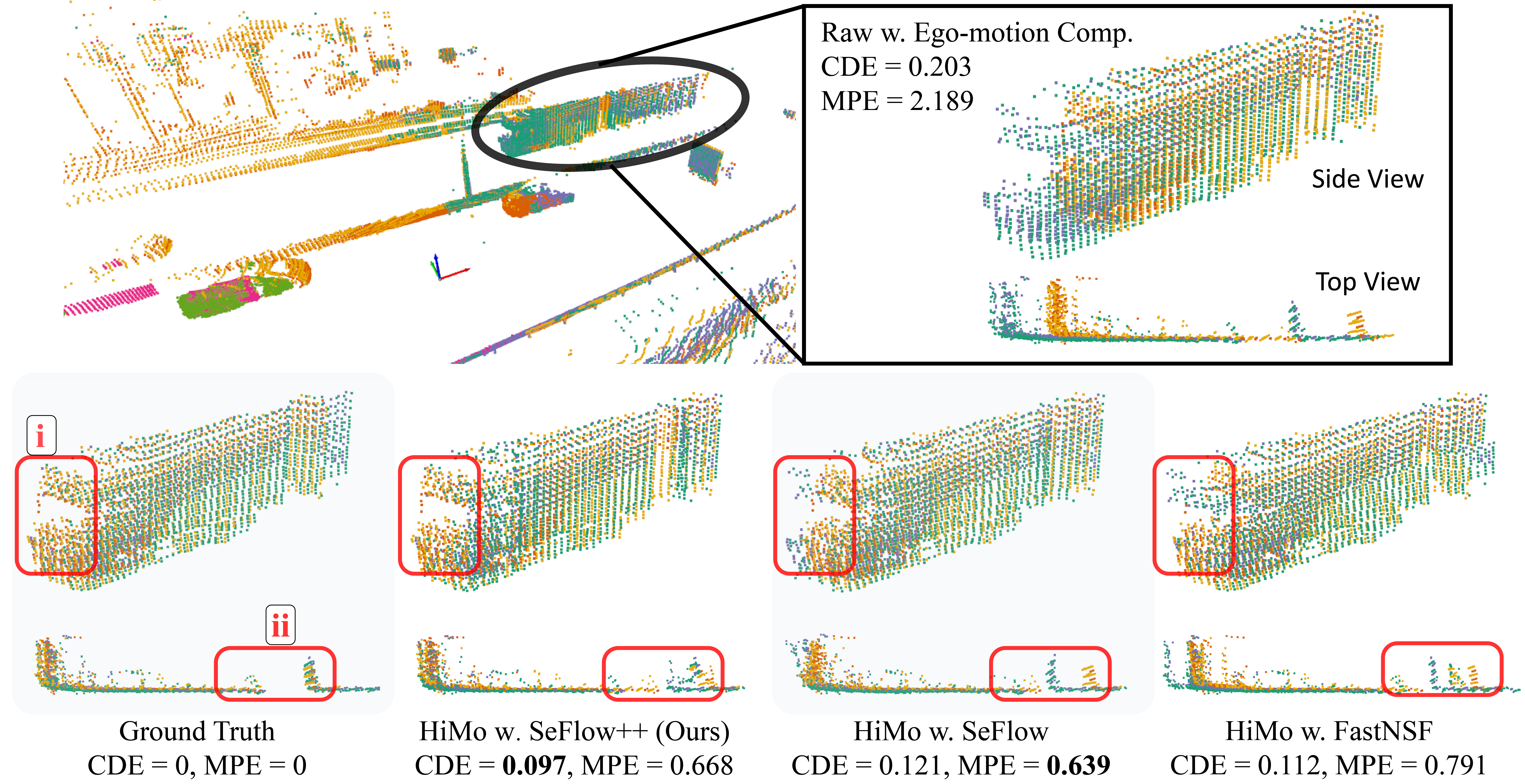}
    \caption{Qualitative comparison of scene flow methods inside the HiMo pipeline for motion compensation on a distorted truck object. Top: Raw point cloud data after the baseline ego-motion compensation showing evident residual motion distortion. Bottom (left to right): Ground truth, SeFlow++ (Ours), SeFlow, and FastNSF results. Side and top views are provided for each method. CDE and MPE values are also reported to complement the visualizations.}
    \label{fig:qual_mpe_cde}
    \vspace{-1em}
\end{figure*}

\begin{figure*}
    \centering
    \includegraphics[width=0.99\linewidth]{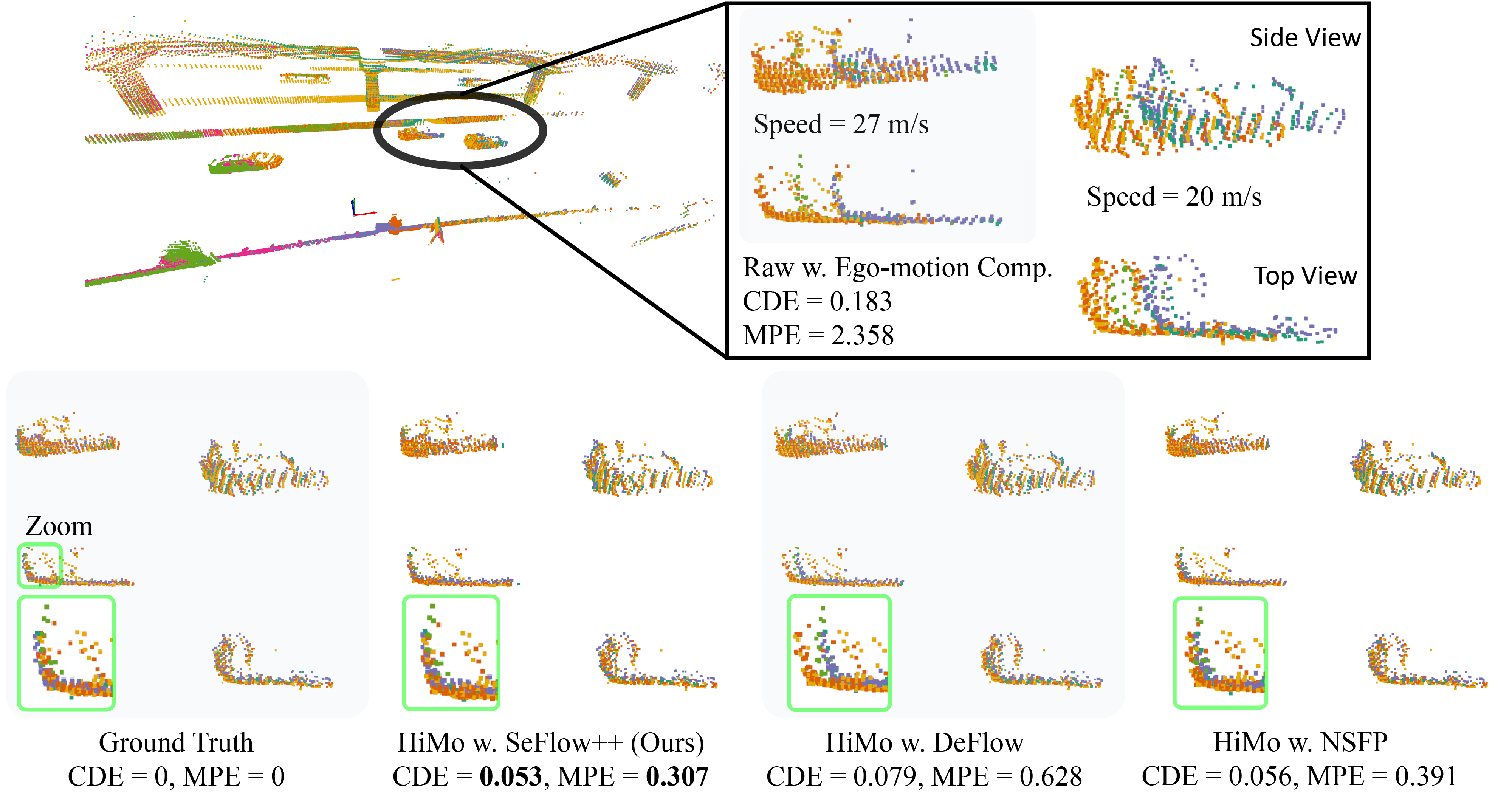}
    \caption{Qualitative comparison of scene flow methods inside the HiMo pipeline for motion compensation on two distorted regular car objects. Top: Raw point cloud data after the baseline ego-motion compensation showing evident distortion. Bottom (left to right): Ground truth, SeFlow++ (Ours), DeFlow, and NSFP results. Side and top views are provided for each method. All scene flow methods in the HiMo pipeline demonstrate effective compensation for the distorted data, with SeFlow++ showing superior performance in both metrics.}
    \label{fig:qualitative_car}
    \vspace{-0.5em}
\end{figure*}

\subsection{Error Distribution}
In \cref{fig:err_object}, we explore the error distributions associated with objects in the CAR category across varying velocities.
The x-axis represents the velocity range of the objects, segmented into intervals (0-10 \si{m/s}, 10-20 \si{m/s}, etc.). 

From the \textit{baseline} ego-motion compensation line plots in \cref{fig:err_object}, it is evident that higher object velocity is directly correlated with larger errors in both CDE and MPE.
This trend confirms our discussions on distortion impacts detailed in \cref{sec:distortion}.

Among the methods, DeFlow, which was trained on Argoverse 2 with ground truth supervision and directly applied to our Scania data, exhibits increased errors as the object velocity increases. This suggests that DeFlow's adaptability to high-speed scenarios is constrained, likely due to its training not fully capturing the fast-moving dynamics. 
The low performance of this pre-trained supervised model on fast-moving objects reveals the necessity of self-supervised learning. 
In comparison, both self-supervised methods (NSFP and SeFlow++) in the HiMo pipeline achieve consistently low CDE and MPE across different object velocities, indicating the advantages of self-supervised learning in handling rapid motion dynamics.

\begin{figure}[t]
\centering
\includegraphics[trim=0 362 1100 5, clip, width=\linewidth]{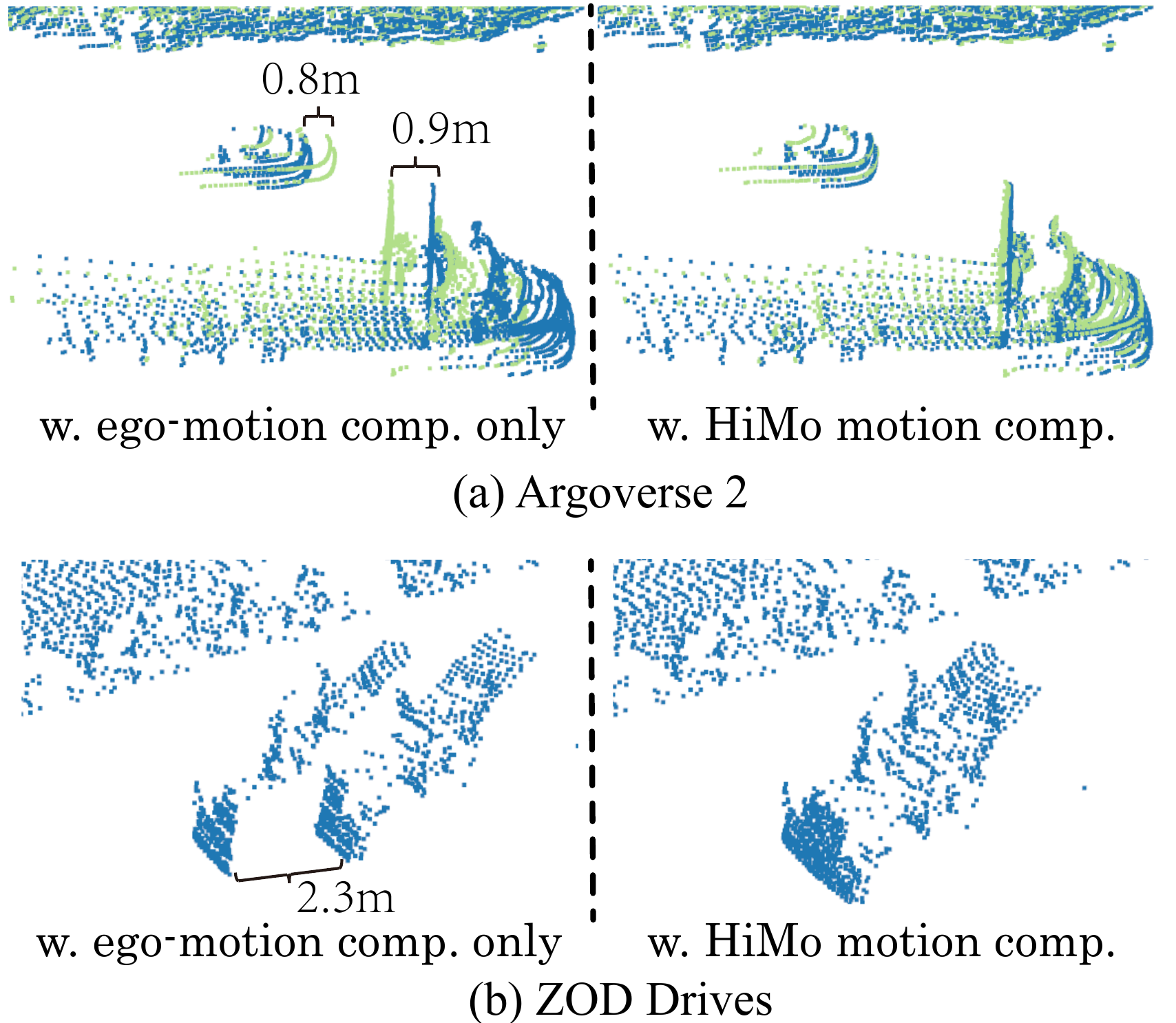}
\caption{
Our HiMo qualitative results on two public datasets. Different point colors in Argoverse 2 represent different source LiDARs. 
The left half of the figure shows the ego-motion compensation results provided by the dataset, and the right side is our HiMo with SeFlow++ undistortion results.
}
\label{fig:qual_others}
\vspace{-1.0em}
\end{figure}

\subsection{Qualitative Results}
Our qualitative analysis, as shown in \cref{fig:qual_mpe_cde}, provides visual results on the performance of different motion compensation methods on a truck object that exhibits significant distortion. 
The ego-motion compensated baseline data demonstrates severe distortion on the truck, with its shape appearing extended and fragmented due to its motion during LiDAR scanning. 
All scene flow estimators in our HiMo pipeline, including SeFlow++, SeFlow, and FastNSF, show improvements over the baseline in terms of both quantitative metrics (CDE and MPE) and visual appearance. 
However, each method exhibits different aspects in its refinement results.
FastNSF provides a balanced performance, improving over the baseline in both CDE and MPE, but does not match the refinement quality of SeFlow in point accuracy or SeFlow++ in shape reconstruction.
SeFlow demonstrates the best performance in terms of MPE, indicating high point-level accuracy. 
However, this superiority in point accuracy is not immediately apparent in the visual representation. 
After applying HiMo with SeFlow++, the truck's outline and structure are more coherent and closely resemble the ground truth (\cref{fig:qual_mpe_cde}.i), as evidenced by its lowest CDE among all evaluated methods.
Interestingly, SeFlow++ shows a more scattered point distribution at the center part of the truck (\cref{fig:qual_mpe_cde}.ii), which may explain its slightly higher MPE despite better shape preservation.
These observations highlight the importance of considering both shape similarity (CDE) and point-level accuracy (MPE) in evaluating motion compensation methods. 
While an ideal method would excel in both metrics, practical limitations often lead to tradeoffs. 
Among the evaluated approaches in the HiMo pipeline, SeFlow++ most effectively preserves object-level geometric integrity, and it is particularly critical for downstream perception tasks in autonomous driving.

Another qualitative motion compensation result on two regular car objects is presented in \cref{fig:qualitative_car}.
All methods in our HiMo demonstrate significant improvements in both CDE and MPE for cars against the baseline, with more substantial reductions compared to the truck scenario. 
For instance, SeFlow++ achieves a remarkable 71\% reduction in CDE  and an 89\% decrease in MPE. 
DeFlow, despite being trained on different datasets, shows competitive performance in CDE reduction (57\% decrease to 0.079) for the two CAR-type objects. 
However, its compensation for the faster-moving car (the left vehicle) with the speed of 27 \si{m/s} appears less refined compared to SeFlow++ and NSFP, particularly in preserving the vehicle's shape integrity. 

The qualitative results in \cref{fig:qual_mpe_cde} and \cref{fig:qualitative_car} on the Scania dataset of other flow methods tested in the HiMo pipeline can be found on the project page \url{https://kin-zhang.github.io/HiMo}.
To demonstrate the generalizability of our HiMo pipeline, we provide qualitative motion compensation results on public datasets Argoverse 2~\cite{Argoverse2} and ZOD~\cite{alibeigi2023zenseact} in \cref{fig:qual_others}.

\begin{table}
\centering
\def\arraystretch{1.3}
\caption{
Comparative analysis of undistortion performance and total computational time. 
Performance is the average reduction in error percentage on CDE and MPE compared to baseline ego-motion compensated data, illustrating the undistortion performance efficacy of each method in HiMo.
Computational Time (hours) contains training time and undistortion inference time on the full Scania dataset (around 60,000 frames).
}
\begin{tabular}{lcccc} 
\toprule
\multirow{2}{*}{\begin{tabular}[c]{@{}c@{}}HiMo w.\\Flow Estimator\end{tabular}} & \multirow{2}{*}{Performance ↑} & \multicolumn{3}{c}{Computational Time (hours) ↓}  \\
&                              & Training   & Inference    & Total            \\ 
\midrule
NSFP \cite{li2021neural}        & \underline{73.51\%}   & \textbf{0} & 250          & 250              \\
FastNSF \cite{li2023fast}       & 71.35\%               & \textbf{0} & 83           & 83               \\
ICP-Flow \cite{lin2024icp}      & 30.62\%               & \textbf{0} & 150          & 150              \\
SeFlow \cite{zhang2024seflow}   & 58.93\%               & 12         & \textbf{0.8} & \textbf{13}      \\
SeFlow++ (Ours)                 & \textbf{76.21\%}      & 14         & \underline{1}    & \underline{15}       \\
\bottomrule
\end{tabular}
\label{tab:runtime}
\vspace{-0.5em}
\end{table}

\begin{figure*}[t]
\centering
\includegraphics[trim=0 20 0 0, clip,width=0.98\textwidth]{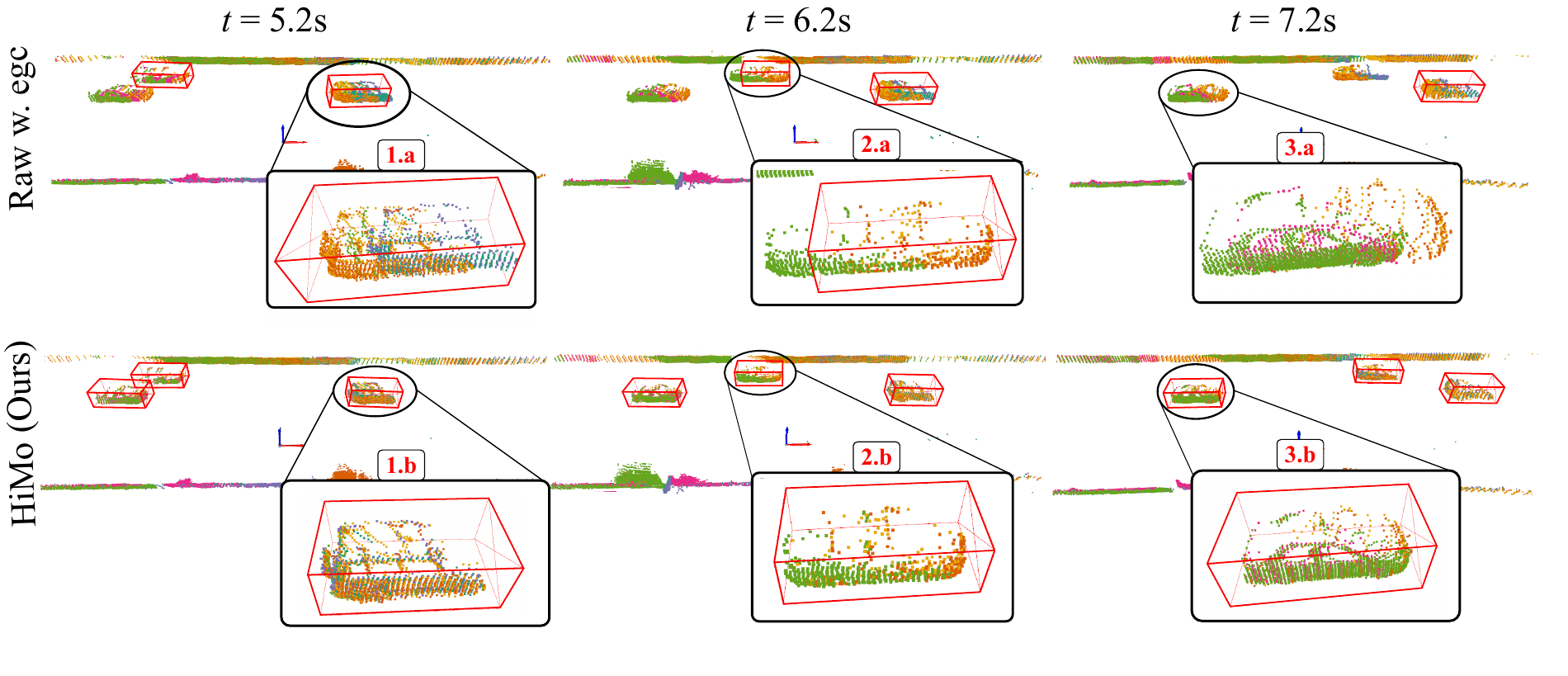}
\caption{
Impact of motion distortion on 3D detection using TransFusion~\cite{bai2022transfusion} applied to Scania highway scenes. 
The examples are from the same scene at different timestamps during a 2-second sequence ($t=5.2$, $6.2$, $7.2$\,s). 
Top: detection results on data with ego-motion compensation show three different failure modes. (1.a) shows an over-extended bounding box with incorrect orientation; (2.a) shows missing geometry due to distortion; (3.a) shows a failure in detection (no bounding box).
Bottom: detection results on undistorted data processed by the HiMo pipeline. Note how the proposed data undistortion mitigates all three detection errors over the same 2s sequence. 
As shown in the figure, detection using HiMo-processed data produces compact, accurate, and consistent bounding boxes.
}
\label{fig:detection_qual}
\vspace{-1em}
\end{figure*}

\subsection{Computational Cost Comparison}
\label{sec:runtime}
In this section, we consider the computational cost for the three top-performing self-supervised methods in our HiMo pipeline: SeFlow++, NSFP, and FastNSF as shown in \cref{tab:main_res}. 
These methods differ significantly in training duration, inference time and deployment efficiency, all of which are crucial factors for practical applications.

As shown in \cref{tab:runtime}, the computational time of SeFlow++ can be decomposed into 4 hours of preparation time (auto-labeling), 10 GPU hours of training, and 1 hour of inference to undistort the whole Scania dataset (around 60,000 frames).
Despite its considerable initial cost during the auto-labeling and training stage, SeFlow++ excels in inference and deployment efficiency afterward on large datasets. 
It requires less than one GPU hour to perform motion compensation on the entire Scania dataset (0.06 seconds per frame). 
Conversely, NSFP, the second top-performing method on the Scania dataset, demands substantially more computational resources. 
It requires approximately 250 GPU hours to apply undistortion across all frames of the full dataset, averaging 15 seconds per frame. 
FastNSF, while quicker than NSFP, still requires around 83 GPU hours (3 seconds per frame). 
However, both NSFP and FastNSF are runtime optimization methods that do not require prior training. As such, they allow for quick deployment and are advantageous if one only needs to perform motion compensation on a handful of frames.

Nevertheless, given the increasing amount of data and the need for online undistortion in future scenarios, HiMo with SeFlow++ emerges as more advantageous in terms of runtime. Its deployment speed is nearly 100 times faster than that of NSFP and 20 times faster than FastNSF. 

\subsection{Downstream Effects}
Our proposed HiMo pipeline is designed to correct motion-related geometric distortions in LiDAR point clouds. These distortions impact subsequent downstream tasks such as scene understanding, semantic segmentation, 3D detection and decision planning. 
As shown in~\cref{fig:qual_others}, distortion can lead to visibly incorrect object shapes, duplicated contours and fragmented structures. These artifacts are not only misleading to human annotators but also degrade the performance of perception models, especially in high-speed driving scenarios.

\noindent\textbf{Scene Understanding} 
To quantify the impact of motion distortion on scene understanding, we evaluate the HiMo motion compensation method on the recent segmentation model WaffleIron~\cite{puy23waffleiron}, chosen due to its competitive performance and ease of deployment.
The model is trained on KITTI - an urban dataset with mainly low-speed scenes and therefore low distortion. 
We then apply WaffleIron to the high-speed Argoverse 2 validation frames, comparing two input variants: (i) raw point clouds with baseline ego-motion compensation and (ii) point clouds corrected by HiMo.
To ensure a fair evaluation across different sensor setups and annotation coverage, 
we report two sets of IoU scores. ``All'' reports the IoU over all points, whilst ``Mask only'' reports the IoU over points that fall into the labelled ground-truth bounding boxes.
As shown in~\Cref{tab:seg}, HiMo consistently improves segmentation performance across both the CAR and OTHERS categories.
In the ``Mask Only'' setting, the IoU for OTHERS increases by 3.95 points — a relative gain of over 12.5\% — despite no fine-tuning of the segmentation model. 
These results demonstrate that HiMo improves object representation accuracy by preventing the segmentation model from misinterpreting fast-moving objects (e.g. cars), which might otherwise appear elongated due to motion distortions.
It shows that even without fine-tuning, existing models benefit from inference on distortion-corrected data, highlighting the semantic consistency achieved through our correction.

\begin{table}[h]
\centering
\caption{
Segmentation IoU (\%) on high-speed Argoverse 2 validation frames using WaffleIron~\cite{puy23waffleiron}.
We compare segmentations of raw point clouds with ego-motion compensation and point clouds corrected by HiMo. Results are reported for the CAR and OTHERS categories over all points and ground-truth-labeled regions only.
}
\def\arraystretch{1.3}
\begin{tabular}{ccccc} 
\toprule
\multirow{2}{*}{Input Point Cloud}       & \multicolumn{2}{c}{Mask Only} & \multicolumn{2}{c}{All}  \\ 
\cmidrule{2-5}
                                         & CAR    & OTHERS               & CAR    & OTHERS          \\ 
\midrule
\multicolumn{1}{l}{w. Ego-motion Comp.}  & 80.90 & 31.44                & 66.08 & 9.83           \\
\multicolumn{1}{l}{w. HiMo Motion Comp.} & 81.51 & 35.39               & 66.43 & 11.15           \\
\bottomrule
\end{tabular}
\label{tab:seg}
\vspace{-0.5em}
\end{table}

\noindent\textbf{3D Detection}
We further analyze the impact of distortion on 3D object detection by applying TransFusion~\cite{bai2022transfusion}, a well-performing detector with publicly available NuScenes-pretrained weights. 
We use the implementation from OpenPCDet~\cite{openpcdet2020}, making it easy to swap in other detectors in the codebase if desired.
Again, we compare the object detection results using two inputs: the raw point clouds with baseline ego-motion compensation, and the undistorted point clouds produced by the HiMo pipeline. 
The comparison is shown in~\Cref{fig:detection_qual} on the Scania dataset.
Using the ego-motion compensated baseline point cloud as input (a), the pretrained detector exhibits three different failure modes even within a short 2-second sequence.
This highlights how non-ego motion distortions in point clouds can cause significant degradation in detection quality. 
In contrast, with HiMo-corrected data as input (b), the same detector produces consistently aligned, compact, and correctly oriented bounding boxes, mitigating multiple downstream errors within this brief 2s sequence.
These examples highlight how motion-induced distortions degrade the reliability of perception, especially in high-speed and multi-LiDAR settings, while HiMo offers a direct and effective correction strategy.

\noindent\textbf{Decision-making and Planning}
Above, we have shown qualitative and quantitative results of HiMo compensation on downstream perception tasks. 
The errors introduced by these motion distortions, such as incorrect object shapes and duplicated positions, will also propagate into downstream decision-making and planning tasks that rely on correct 3D representations. These errors may lead to obstacle avoidance failures, unstable trajectory planning, or delayed decision-making. 
By restoring object geometry and motion consistency, HiMo can also enhance the reliability of planning systems built on top of it.

\section{Conclusion}
In this paper, we addressed the critical issue of motion-induced distortions in LiDAR point clouds, which significantly impact the accuracy of environmental perception in autonomous driving systems. 
We analyzed the source of distortions and found that in addition to ego-motion, the motion of surrounding objects was a large source of distortion.
Our investigation revealed the existence of moving object distortions across various datasets, particularly affecting high-speed objects and multi-LiDAR setups.

To tackle this challenge, we introduced HiMo, a novel pipeline that repurposes scene flow estimation for non-ego motion compensation. 
To further enhance motion estimation within HiMo, we also propose SeFlow++, a real-time self-supervised scene flow estimator that refines dynamic classification and symmetric loss, improving training efficiency and performance with smaller datasets.
We demonstrated the effectiveness of HiMo through extensive experiments on our Scania highway and public datasets.
These experiments compared the undistorted performance of HiMo with the ego-motion compensation baseline and evaluated the impact of different scene flow methods within HiMo. 
The results show that HiMo significantly improves point cloud representations, benefiting downstream tasks such as semantic segmentation and 3D detection, even without any fine-tuning.
Our work contributes to the field by comprehensively analyzing point cloud distortions, proposing an effective compensation method, and offering open-source evaluation data and code. 

Future work could explore the integration of HiMo with developing assistance 3D detection annotation systems as well as real-time perception systems.

\section*{Acknowledgement}
Thanks to Bogdan Timus and Magnus Granström from Scania and Ci Li from KTH RPL, who helped with this work. 
We also thank Yixi Cai, Yuxuan Liu, Peizheng Li and Shenghai Yuan for helpful discussions during revision.
We thank the anonymous reviewers and the associate editor for their useful comments.
The computations were enabled by the supercomputing resource Berzelius provided by National Supercomputer Centre at Linköping University and the Knut and Alice Wallenberg Foundation, Sweden.

\appendix

\subsection{Annotated Ground Truth}
\label{sec:anno_gt}
\begin{figure}
\centering
\includegraphics[width=0.7\linewidth]{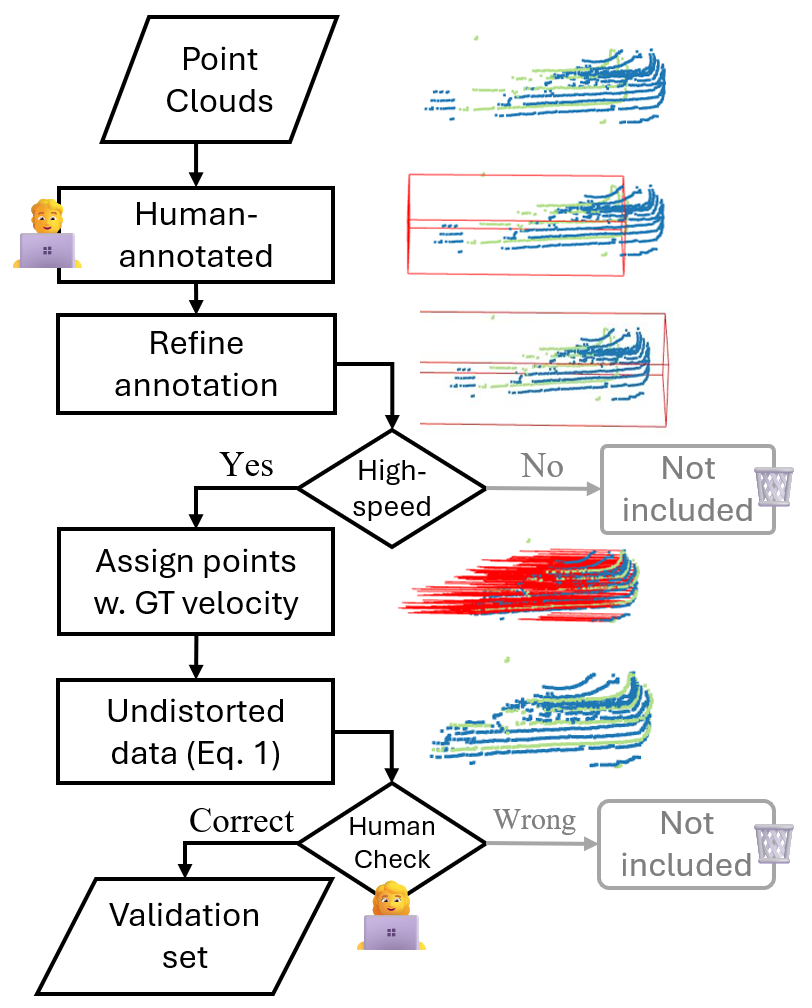}
\caption{
Flowchart of the annotated ground-truth generation process. 
The process starts with manual annotation and velocity-based refinement of object bounding boxes. The velocities of these bounding boxes between frames are then used to undistort the raw data. Finally, we manually verify the ground-truth before including it in the validation set.
}
\label{fig:gt_anno}
\vspace{-1.0em}
\end{figure}

\Cref{sec:main_anno_gt} provides a detailed description of the ground-truth generating process for both the Scania and Argoverse 2 datasets. The process is also illustrated in the flowchart in~\cref{fig:gt_anno}.
We also analyze how the annotation strategy, specifically the use of velocity-aware bounding box expansion, affects current scene flow evaluation in~\cref{sec:av2_sf_fixgt}.

\subsection{Scene Flow Performance}
\label{sec:sf_res}
To support our claim that SeFlow++ achieves state-of-the-art performance in real-time self-supervised scene flow estimation, we report its results on the Argoverse 2 public test set leaderboard~\cite{onlineleaderboard}. As shown in~\cref{tab:leaderboard}, SeFlow++ outperforms all other real-time methods, including SeFlow and ZeroFlow, in Three-way EPE mean metrics.

While several optimization-based methods (e.g., NSFP, EulerFlow, Floxels) achieve high accuracy, their runtimes span from several minutes to hours per sequence, making them unsuitable for real-time applications. In contrast, SeFlow++ offers a favorable trade-off between accuracy and efficiency, completing each sequence in just 8.2 seconds.

\begin{table}[h]
\def\arraystretch{1.1}
\centering
\caption{
Scene flow performance comparisons on Argoverse 2 \underline{test set} from the public leaderboard~\cite{onlineleaderboard}.
Our proposed SeFlow++ achieves state-of-the-art performance in real-time self-supervised scene flow estimation. Runtime is reported per sequence (around 157 frames), with `-' indicating unreported runtime. `s', `m', and `h' represent seconds, minutes, and hours, respectively.
{\color{red}Red} highlights the runtime of offline methods. For each method, we report its end point error (EPE) for foreground dynamic (FD), foreground static (FS), background static (BS) points, as well as the average EPE of all three point categories (three-way).
}

\begin{tabular}{lccccc} 
\toprule
\multicolumn{1}{c}{\multirow{2}{*}{Methods}} & \multirow{2}{*}{\begin{tabular}[c]{@{}c@{}}Runtime\\per seq\end{tabular}} & \multicolumn{4}{c}{EPE (cm) ↓}  \\
\multicolumn{1}{c}{}                         &                                                                           & Three-way  & FD    & FS   & BS                 \\ 
\midrule
Ego Motion Flow                              & -                                                                         & 18.13 & 53.35 & 1.03 & 0.00               \\ 
\midrule
FastNSF~\cite{li2023fast}                                      & \textcolor{red}{12m}                                                                       & 11.18 & 16.34 & 8.14 & 9.07               \\
NSFP~\cite{li2021neural}                                         & \textcolor{red}{1.0h}                                                                      & 6.06  & 11.58 & 3.16 & 3.44               \\
Floxels~\cite{hoffmann2025floxels}                                      & \textcolor{red}{24m}                                                                       & \textbf{3.57}  & 7.73  & \textbf{1.44} & \textbf{1.54}               \\
EulerFlow~\cite{vedder2024neural}                                    & \textcolor{red}{24h}                                                                      & 4.23  & \textbf{4.98}  & 2.45 & 5.25               \\ 
\midrule
ZeroFlow~\cite{zeroflow}                                     & 5.4s                                                                      & 4.94  & 11.77 & 1.74 & 1.31               \\
ICP-Flow~\cite{lin2024icp}                                     & -                                                                         & 6.50  & 13.69 & 3.32 & 2.50               \\
SemanticFlow~\cite{chen2025semanticflow}                                 & -                                                                         & 4.69  & 12.26 & \textbf{1.41} & \textbf{0.40}               \\
SeFlow~\cite{zhang2024seflow}                                       & 7.2s                                                                      & 4.86  & 12.14 & 1.84 & 0.60               \\
VoteFlow~\cite{lin2025voteflow}                                       & 8.0s                                                                      & 4.60  & 12.14 & 1.84 & 0.60               \\
SeFlow++ (Ours)                                     & 8.2s                                                                      & \textbf{4.40}  & \textbf{10.99} & 1.44 & 0.79               \\
\bottomrule
\end{tabular}
\label{tab:leaderboard}
\vspace{-1em}
\end{table}

\subsection{Motion Compensation in Scene Flow Ground Truth}
\label{sec:av2_sf_fixgt}
Originally, the Argoverse 2 dataset expands scene flow human-annotated bounding boxes by a fixed 20\si{cm} offset to account for distorted points~\cite{Argoverse,chodosh2024re}. 
However, this method is ineffective in high-speed scenarios where the motion distortion can be significantly larger than 20\si{cm}. As a result, the labelled bounding boxes might not completely cover the entire object, leading to errors in the 3D scene flow labels. 
To mitigate such errors, following our HiMo pipeline insight, we expand the bounding boxes based on the relative velocity of dynamic objects, ensuring all relevant points are included. 

\Cref{tab:fixed_gt} shows the effect of this correction of groundtruth labels\footnote{More detailed information can be found in \url{https://github.com/KTH-RPL/OpenSceneFlow/pull/5} with visualization included.}.
Compared to the GT corrected with fixed offset (20\si{cm}), all methods perform worse when evaluated using the GT corrected with our velocity-based procedure. 
The drop in performance is most visible in the FD category that captures errors in foreground dynamic points. 
This indicates that the previous ground truth fails to properly capture the dynamics of all foreground objects, especially those with high speed, leading to underestimated flow errors.

\begin{table}
\centering
\caption{
Impact of ground truth correction process on the performance of three scene flow estimators. The original fixed bounding box expansion procedure (`Fixed (20cm)') led to missed points and inflated scores, particularly for dynamic objects. 
Our updated strategy with velocity-based bounding box expansion (`Velocity-based') ensures that all points belonging to the object are included, with flow values assigned, correcting the previously inflated scores.
Pre-trained weights, as released by the original papers, are used for this evaluation.
}
\def\arraystretch{1.2}
\begin{tabular}{cccccc} 
\toprule
\multirow{2}{*}{Method} & \multirow{2}{*}{GT Correction}              & \multicolumn{4}{c}{Three-way EPE (cm) ↓}                                                                                                                        \\ 
\cline{3-6}
                        &                                        & Mean                                  & FD                                     & FS                                    & BS                                     \\ 
\midrule
\multirow{2}{*}{Flow4D~\cite{kim2024flow4d}} & Fixed (20cm)                                 & 3.09                                  & 6.98                                   & 1.17                                  & 1.11                                   \\
                        & {\cellcolor[rgb]{0.91,0.91,0.91}}Velocity-based & {\cellcolor[rgb]{0.91,0.91,0.91}}3.80 & {\cellcolor[rgb]{0.91,0.91,0.91}}8.88  & {\cellcolor[rgb]{0.91,0.91,0.91}}1.66 & {\cellcolor[rgb]{0.91,0.91,0.91}}0.86  \\ 
\midrule
\multirow{2}{*}{DeFlow~\cite{zhang2024deflow}} & Fixed (20cm)                                 & 3.40                                  & 7.21                                   & 2.00                                  & 1.00                                   \\
                        & {\cellcolor[rgb]{0.91,0.91,0.91}}Velocity-based & {\cellcolor[rgb]{0.91,0.91,0.91}}3.93 & {\cellcolor[rgb]{0.91,0.91,0.91}}9.12  & {\cellcolor[rgb]{0.91,0.91,0.91}}1.94 & {\cellcolor[rgb]{0.91,0.91,0.91}}0.74  \\ 
\midrule
\multirow{2}{*}{SeFlow~\cite{zhang2024seflow}} & Fixed (20cm)                                 & 5.84                                  & 13.83                                  & 2.24                                  & 1.45                                   \\
                        & {\cellcolor[rgb]{0.91,0.91,0.91}}Velocity-based & {\cellcolor[rgb]{0.91,0.91,0.91}}6.29 & {\cellcolor[rgb]{0.91,0.91,0.91}}15.56 & {\cellcolor[rgb]{0.91,0.91,0.91}}2.16 & {\cellcolor[rgb]{0.91,0.91,0.91}}1.16  \\
\bottomrule
\end{tabular}
\label{tab:fixed_gt}
\vspace{-1em}
\end{table}

{
\small
\bibliographystyle{IEEEtran}
\bibliography{IEEEabrv,ref}
}

\end{document}